\documentclass[10pt,twocolumn,letterpaper]{article}

\usepackage{wacv}
\usepackage{times}
\usepackage{epsfig}
\usepackage{graphicx}
\usepackage{amsmath}
\usepackage{amssymb}
\usepackage{caption}
\usepackage{array,multirow,booktabs}
\usepackage{hyperref}
\usepackage{capt-of,etoolbox}
\usepackage{amsfonts}

\usepackage{url}
\usepackage[table]{xcolor}
\definecolor{newcolor}{rgb}{.8,.349,.1}
\usepackage[shortlabels]{enumitem}
\usepackage{multirow}
\usepackage{booktabs}

\wacvfinalcopy 

\newcommand{\algoname}{DIHE}
\newcommand{\extendedname}{Domain invariant hierarchical embedding}

\begin{document}

\title{\extendedname{} for grocery products recognition}

\author{Alessio Tonioni\\
	DISI, University of Bologna\\
	{\tt\small alessio.tonioni@unibo.it}
	\and
	Luigi Di Stefano\\
	DISI, University of Bologna\\
	{\tt\small luigi.distefano@unibo.it}
}

\maketitle
\ifwacvfinal\thispagestyle{empty}\fi

\begin{abstract}
Recognizing packaged grocery products based solely on appearance is still an open issue for modern computer vision systems due to peculiar challenges. Firstly, the number of different items to be recognized is huge (\ie, in the order of thousands) and rapidly changing over time. Moreover, there exist a significant domain shift between the images that should be recognized at test time, taken in stores by cheap cameras, and those available for training, usually just one or a few studio-quality images per product. 
We propose an end-to-end architecture comprising a GAN to address the domain shift at training time and a deep CNN trained on the samples generated by the GAN to learn an embedding of product images that enforces a hierarchy between product categories. At test time, we perform recognition by means of K-NN search against a database consisting of just one reference image per product. Experiments addressing recognition of products present in the training datasets as well as different ones unseen at training time show that our approach compares favourably to state-of-the-art methods on the grocery recognition task and generalize fairly well to similar ones.
\end{abstract}


\section{Introduction}
\label{sec:introduction}
Automatic recognition of grocery products is receiving increasing attention as it may lead to improved shopping experience (\eg, shopping apps, interaction via augmented reality, checkout-free stores, support to visually impaired customers) as well as to a more efficient store management (\eg, by automated inventory and on-line shelf monitoring). 

\begin{figure*}
	\centering
	\begin{tabular}{cccc}
		\includegraphics[height=4cm]{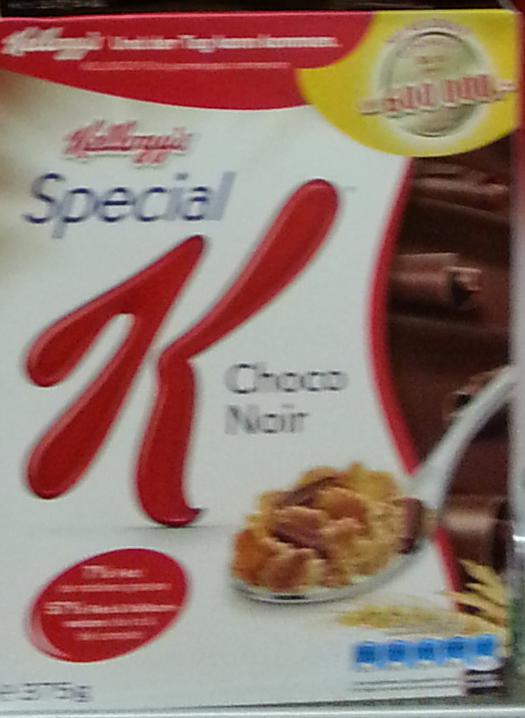} & 
		\includegraphics[height=4cm]{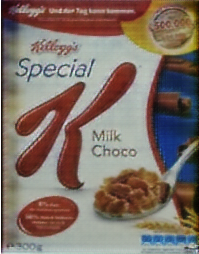} & 
		\includegraphics[height=4cm]{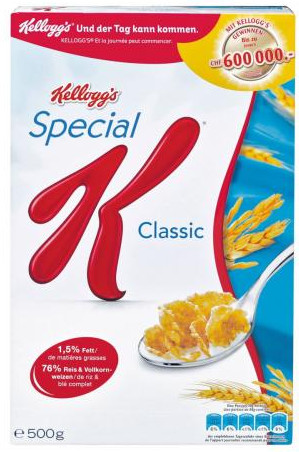} &
		\includegraphics[height=4cm]{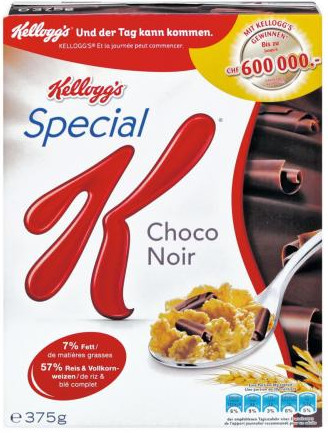} \\
		(a) & (b) & (c) & (d)\\		
	\end{tabular}
	\caption{Exemplar images for the grocery recognition task: \textit{reference} images (c-d) carefully acquired in studio (available at training time), \textit{query} images: (a) captured in the store (\textit{query} at test time) and (b) \textit{synthetic query} generated by our GAN (used at training time).}
	\label{fig:examples}
\end{figure*}

As pointed out by \cite{merler2007recognizing} in their seminal work, recognizing grocery products can be thought of as an object recognition problem featuring peculiar challenges. Firstly, the number of different items to be recognized is huge, in the order of several thousands for small to medium shops, well beyond the usual target for current state-of-the-art image classifiers. Furthermore, product recognition is better cast as an hard instance recognition rather than a classification problem, as it mandates telling apart many items looking identical but for a few details, such as the different flavours of the same cereal brand depicted in \autoref{fig:examples}(b-c-d). Any practical methodology should rely only on the model images available within existing commercial product databases, \ie{} a single high-quality image for each side of the package either acquired in studio settings or rendered using computer graphics tools (\eg, \autoref{fig:examples}(c-d)). Conversely, products must be recognized from images captured in the store by cheap sensors, \eg{,} smartphone cameras, under far less than ideal conditions (\eg, \autoref{fig:examples}-a). Thus, product recognition implies tackling a domain adaptation problem between the images available to build the inference engine and those deployed at test time to pursue recognition. 
Another peculiarity deals with the items on sale in a store as well as their appearance changing frequently overtime, which would render unfeasible to continuously re-train or fine-tune an inference engine in order to keep-up with all such changes. Differently, a practical approach should be conducive to recognition of both \emph{seen products} (\ie, products whose reference images were deployed at training time) as well as \emph{unseen products} (\ie, products present in the store but not used to train the inference engine). 

Given these premises, we rely on a global image descriptor learned to disentangle grocery products and to pursue recognition through K-NN search within a database featuring one reference image per sought product. This approach allows for learning an image embedding based on the available training data and then perform recognition of both \emph{seen} and \emph{unseen} products seamlessly. For example, should a new product be put on sale in the store, our system would  just require to add one image of the new product into the reference database without the need of performing a new costly retraining. K-NN search is quite amenable to product recognition from a computational standpoint alike. Indeed, compared to typical image retrieval settings, the database is very small (\ie{,} in the order of several thousands images rather than millions of images). Thus, a global image descriptor of about a few hundreds entries turns out viable in terms of time and memory efficiency. 

We propose to learn the embedding through a deep CNN trained by a loss function that forces both similar looking items as well as items belonging to the same high level category to map close one to another in the descriptor space. Moreover, to tackle the domain shift issue and increase the training set size we propose to deploy an image-to-image translation GAN together with the embedding CNN and to optimize the whole architecture end-to-end. In particular, some of the training samples for the embedding network are generated by a GAN that learns unsupervisedly to transform images taken in studio settings into in-store type of images without introducing excessive modification of product appearance (\autoref{fig:examples} (b) shows an exemplar image synthesized by the GAN). These training samples force the embedding CNN to learn robustness to domain shift; moreover, the GAN can be trained to produce samples that are particularly hard to embed, thereby allowing the CNN to learn a stronger embedding function thanks to these adversarial samples. Despite the use of multiple networks, the overall architecture can be trained end-to-end effortlessly via simple gradient descent.

Thus, the main original contributions of this paper can be summarized as follows. 
\begin{itemize}
	\item The introduction of an image-to-image translation GAN trained jointly together with an embedding network in order to produce a domain-shift resilient and stronger embedding function. To the best of our knowledge, this is the first work that tries to integrate a GAN with adversarial behaviour within the training process of an embedding network. 
	\item A novel formulation of the classic triplet ranking loss that can be used to learn a better embedding whenever in the domain of interest there exist a taxonomy between classes  (\eg, ImageNet class taxonomy). Our loss helps preserving in the descriptor space the similarity information implied by the taxonomy  (\ie, items belonging to the same macro class should be embedded closer than those belonging to unrelated classes).
	\item A novel formulation of the grocery product recognition task as an instance-level recognition problem with thousands of classes and only one sample per class. In \autoref{sec:experimental results} we will show how in this real-world scenario, far more challenging than the standard datasets currently used to benchmark methods aimed at learning from few shots, our proposed architecture can obtain impressive performance. 
\end{itemize}


\section{Related Work}

\paragraph{\textbf{Grocery Product Recognition}}
The grocery products recognition problem was firstly investigated in the seminal paper by \cite{merler2007recognizing}. Together with a thoughtful analysis of the problem, the authors propose a system based on local invariant features to help visually impaired costumers shop in grocery stores. However, their experiments report performance far from conducive to real-world deployment.  A number of more recent proposals are aimed at improving product recognition by leveraging on: a) stronger features followed by classification (\cite{cotter2014hardware}), b) the statistical correlation between nearby products on shelves (\cite{advani2015visual,baz2016context}) c) information on the expected product disposition  (\cite{tonioni2017product}) or d) a hierarchical multi-stage recognition pipeline (\cite{franco2017grocery}). Yet, all these recent works focus on a relatively small-scale problem, \ie{} recognition of a few hundreds different items at most, whilst several thousands products are on sale in a real shop. \cite{george2014recognizing} address more realistic settings and propose a quite complex multi-stage system capable of recognizing $\sim3400$ different products based on a single model image per product. The authors contribution include releasing the dataset employed in their experiments, which we will use in our evaluation. Recently \cite{karlinsky2017fine} have shown how it is possible to improve detection and recognition performance on the same dataset relying on a probabilistic model based on local feature matching and refinement by deep network. However, even in this work, performance appear not as satisfactory as to pave the way for practical exploitation.

\paragraph{\textbf{Computer Vision for Retail}}
The recognition of grocery products shares commonalities with the \textit{exact street to shop} task addressed in  \cite{hadi2015buy,wang2016matching,shankar2017deep}, which consists in recognizing a real-world example of a garment item based on the catalog of an online shop. Like our solution, these works rely on matching and retrieval using deep features extracted from CNN architectures. However, they leverage on labeled paired couple of samples depicting the same item in the \textit{Street} and \textit{Shop} domain to learn a cross domain embedding at training time while our proposal leverages only on labeled images from one domain, thereby vastly relaxing the applicability constraint.
Moreover, computer vision has been successfully applied in the retail environments for costumer profiling (\cite{sturari2016robust}), automatic shelf surveying (\cite{paolanti2017mobile}), visual market basket analysis (\cite{santarcangelo2018market}) and automatic localization inside the store (\cite{wang2015lost}).


\paragraph{\textbf{Embedding Learning}}
Using CNNs to obtain rich image representations is nowadays an established approach to pursue image retrieval, both as a strong off-the-shelf baseline (\cite{sharif2014cnn}) and as a key component within more complex pipelines (\cite{gordo2016deep}). \cite{schroff2015facenet} train a CNN using triplets of samples to create an embedding for face recognition and clustering. Since then, this approach has been used extensively to learn representations for a variety of different tasks, with more recent works advocating smart sampling strategies (\cite{wu2017sampling}) or suitable regularizations  (\cite{zhang2017learning}) to ameliorate performance.  Similarly to our proposal, \cite{zhang2016embedding} extend the idea of triplets by a novel formulation amenable to embed label structure and semantics at training time based on tuplets. Unlike \cite{zhang2016embedding,schroff2015facenet}, in this paper we propose to embed label structure within the learning process using only standard triplets; moreover our method uses only one exemplar image per class and augment the training set by a GAN trained jointly together with the embedding network.   

\paragraph{\textbf{Few Shot Learning}}
Few shot learning has been addressed successfully by \cite{hariharan2017low} through classifiers trained on top of a fixed feature representation by artificially augmenting a small training set with transformations in the feature space. Yet, in the grocery product recognition scenario the items to be recognized at test time change quite frequently, which would mandate frequent retraining of new classifiers. Besides, as product packages exhibit very low intra-class variability, the generalization ability of a classifier may not be needed. Thus, we prefer to learn a strong image embedding and rely on K-NN similarly to perform recognition. In this respect our approach shares commonalities with \cite{vinyals2016matching} and \cite{snell2017prototypical}, where the authors address few shot learning by learning suitable embedding spaces and matching functions.  

\paragraph{\textbf{GANs}}
Starting from the pioneering works of \cite{goodfellow2014generative} and \cite{radford2016unsupervised}, GANs have received ever increasing attention in the computer vision community as they enable to synthesize realistic images with few supervision. Recently GAN frameworks have been successfully deployed to accomplish image-to-image translation, with (\cite{isola2016pix}) and without (\cite{zhu2017unpaired,shrivastava2016learning}) direct supervision, as well as to tackle domain shift issues by forcing a classifier to learn invariant features  \cite{tzeng2017adversarial}. We draw inspiration from these works and deploy a GAN at training time to pursue domain adaptation as well as to improve the effectiveness of the learned embedding. A related idea is proposed in \cite{qiu2017deep} though,  unlike \cite{qiu2017deep}, (a) we explicitly deploy the GAN while learning the embedding to attain domain adaptation, (b) use only one sample per class and (c) train the GAN to produce realistic though hard to embed training samples, \ie{} the \textit{generator} of our GAN not only plays an adversarial game against the \textit{discriminator} but also against the \textit{encoder} network that learns the embedding..

\section{\extendedname}
\label{sec:metodo}

\begin{figure*}
	\centering
	\includegraphics[width=0.85\textwidth]{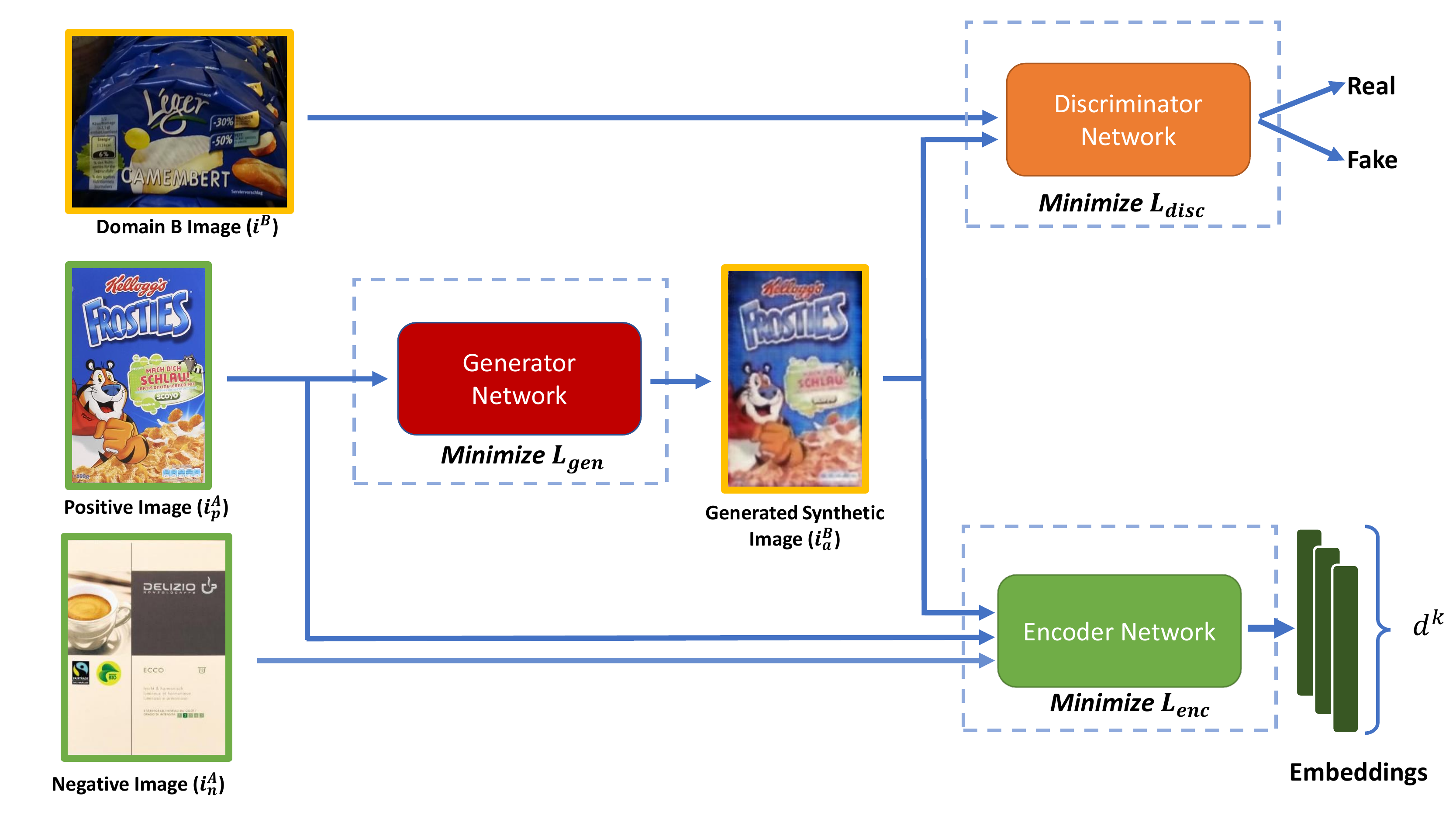}
	\caption{Overview of \algoname{} at training time. Each training sample consists of three images, two from domain $\mathcal{A}$ (enclosed in green) and one from domain $\mathcal{B}$ (enclosed in yellow). The \textit{generator} and \textit{discriminator} implement a classic GAN for domain translation from $\mathcal{A}$ to $\mathcal{B}$. The \textit{encoder} network uses two images from domain $\mathcal{A}$ alongside with the generated one to learn an image embedding by a modified triplet ranking loss. $i_a^\mathcal{B}$ is generated to be both indistinguishable from images sampled from domain $\mathcal{B}$ as well as hard to encode.}
	\label{fig:overview}
\end{figure*}

An overview of our \extendedname{} (\algoname{}) is depicted in \autoref{fig:overview}. We use a deep CNN (\textit{encoder}) to learn an embedding function $E:\mathcal{I}\rightarrow\mathcal{D}$ that maps an input image $i \in \mathcal{I}$ to a k-dimensional descriptor $d^k \in \mathcal{D}$ amenable to pursue recognition through K-NN similarity search. During training we exploit, if available in the training dataset, a taxonomy of classes by means of a novel loss function that forces descriptors of different items to be closer if they share some portion of the taxonomy, distant otherwise. To learn a descriptor robust to the domain shift between train and test data, we use an image-to-image translation GAN, consisting of a \textit{generator} and a \textit{discriminator}, which augments the training set with samples similar to those  belonging to the test domain while simultaneously producing hard examples for the embedding network. The three networks can be trained jointly by standard gradient descent in order to minimize the three loss functions described in the following sections. 

\subsection{Hierarchical Embedding}
\label{ss:hi_emb}
An established approach to learn an effective image embedding consists in training a deep CNN according to the \textit{triplet ranking loss} (\cite{wang2014learning}). In the original formulation, each training sample consists of three different images, referred to as \emph{anchor} ($i_a$), \emph{positive} ($i_p$) and \emph{negative} ($i_n$). In a typical classification scenario these images would be chosen such that $i_a$ and $i_p$ depict the same class while $i_n$ belongs to a different one. Given a distance function in the descriptor space, $d(\mathbf{X},\mathbf{Y})$, with $X,Y\in\mathcal{D}$, and denoted as $E(i)$ the descriptor computed by the encoder network for image $i$, the loss to be minimized is defined as
\begin{equation}
\label{eq:tripletLoss}
\mathcal{L}_{enc} = max( 0,d(E(i_a),E(i_p))-d(E(i_a),E(i_n))+\alpha)
\end{equation}
with $\alpha$ a fixed margin to be enforced between the pair of distances.

We modify this formulation for domains that feature a hierarchical structure between classes (\eg, ImageNet classes taxonomy), so as to mimic this structure within the learned descriptor space. This is a quite common scenario for problems where a multi level classification is available. For instance, existing commercial databases of grocery products feature labels both at instance as well as at multiple category levels (\eg, for the product depicted in \autoref{fig:examples} (c) we would have three different classification labels with increasing generality: Kellog's Special K Classic $\rightarrow$Cereal$\rightarrow$Food).  Our aim is to force the network to embed images nearby in the descriptor space not only based on their appearance but also on higher level semantic cues, like those shared between items belonging to the same macro-class. We argue that doing so will help producing a stronger image descriptor and may provide better generalization to products whose reference images are \emph{unseen} at training time. 

Assuming a taxonomy of classes encoded in a tree like structure, we propose to impose a hierarchy in $\mathcal{D}$ by rendering $\alpha$ inversely proportional to the \textit{amount of hierarchy} shared between the classes of $i_a,i_p$ and $i_n$. Considering an image sample $i$ in the training set, we denote with $c$ its fine class (foil level in the taxonomy) and with $\mathcal{H}(i)$ a set of higher level classes (all the parent nodes in the class tree excluding the common root). Using this notation and defining the minimum and maximum margin, $\alpha_{min}$ and $\alpha_{max}$ respectively, our hierarchical margin, $\alpha \in [\alpha_{min},\alpha_{max}]$, can be computed as:
\begin{equation}
\label{eq:alpha}
\alpha=\alpha_{min}+\left(1-\frac{|\mathcal{H}(i_a)\cap\mathcal{H}(i_n)|}{|\mathcal{H}(i_a)|}\right)\cdot(\alpha_{max}-\alpha_{min})
\end{equation} 
where $|\cdot|$ is the cardinality operator for sets. Thus, if $i_a$ and $i_b$ share all the parent nodes $\alpha=\alpha_{min}$,  whilst the margin is proportionally increased until completely disjoint fine classes will produce $\alpha=\alpha_{max}$.

\subsection{Domain Invariance}
\label{ss:dom_inv}
A common trait across many computer vision tasks is that easily available labelled training data (\eg, tagged images published on-line) are usually sampled from a different distribution than the actual test images. Thus, machine learning models directly trained on such samples, such as embedding networks, will typically perform poorly at test time due to domain shift issues. However, annotating samples from the test distribution, even if possible, is usually very expensive and time consuming. We propose to address this problem by dynamically transforming the appearance of the available labelled training images to make them look similar to samples from the unlabeled test images. This transformation is carried out by two CNNs, refereed to in \autoref{fig:overview} as \textit{generator} and \textit{discriminator}, realizing an image-to-image translation GAN which is trained end-to-end together with the embedding network (\textit{encoder}). 

Given two image domains $\mathcal{A},\mathcal{B}\subset \mathcal{I}$ consisting of $i^\mathcal{A} \in \mathcal{A}$ and $i^\mathcal{B} \in \mathcal{B}$, the standard image-to-image GAN framework can be summarized as a \textit{generator} network that tries to learns a generative function $G:\mathcal{A} \rightarrow \mathcal{B}$ by playing a two player min-max game against a \textit{discriminator} network $D:\mathcal{I} \rightarrow \mathbb{R}$ that tries to classify examples either as real images from $\mathcal{B}$ or fake ones produced by $G$. In the following we will denote with $G(i^\mathcal{A}) \in \mathcal{I}$ the output of the \textit{generator} network given the input image $i^\mathcal{A}$ and with $D(i) \in \mathbb{R}$ the output of the \textit{discriminator} network for image  $i$. To generate samples similar to the images from domain $\mathcal{B}$ without drastically changing the appearance of the input image $i^A$, we introduce an additional term in the generator loss function ($\mathcal{L}_{reg}$) that, similarly to the self regularization term deployed in \cite{shrivastava2016learning}, forces $G(i^A)$ to be visually consistent with $i^A$.
In our architecture, $\mathcal{A}$ is the training set while $\mathcal{B}$ is a small set of unlabeled images from the test data distribution. 


Following the notation introduced in \autoref{ss:hi_emb}, during each training iteration of the whole architecture we sample one image $i^B \in \mathcal{B}$ to train the \textit{discriminator} and two from the other domain $i^\mathcal{A}_p,i^\mathcal{A}_n \in \mathcal{A}$ to train the \textit{encoder} and \textit{generator}. As mentioned in \autoref{ss:hi_emb},  the \textit{encoder} needs triplets of samples to compute its loss, so we synthesize the missing image using the \textit{generator} $i^\mathcal{B}_a=G(i^\mathcal{A}_p)$. With this architecture the triplet used to calculate \autoref{eq:tripletLoss}  consists of two images from domain $\mathcal{A}$ and one from the simulated domain $\mathcal{B}$, thereby mimicking the test conditions where the query images to be recognized will came from $\mathcal{B}$ and the reference images to perform K-NN similarity from $\mathcal{A}$.   

The \textit{encoder} is trained to minimize \autoref{eq:tripletLoss} with the margin defined in \autoref{eq:alpha}. The \textit{discriminator} tries to minimize a standard cross entropy loss:
\begin{equation}
\label{eq:cross_entropy}
\mathcal{L}_{disc}=log(D(i^B))+log(1-D(G(i^A_p)))
\end{equation}
while the \textit{generator} minimizes a loss consisting of three terms:
\begin{equation}
\label{eq:gen_loss}
\begin{split}
\mathcal{L}_{gen} = L_{adv}+\lambda_{reg} \cdot L_{reg}+\lambda_{emb} \cdot L_{emb}  \\
L_{adv}=-log(D(G(i^\mathcal{A}_p))) \\
L_{reg}=\phi(i^\mathcal{A}_p,i^\mathcal{B}_a) \\
L_{emb}=-d(E(i^\mathcal{A}_p),E(G(i^\mathcal{A}_p))) \\
\end{split}
\end{equation}
with  $\phi(x,y)$, $x,y \in \mathcal{I}$ a similarity measure between the appearance of image $x$ and $y$, either at pixel level (\eg, mean absolute difference\dots) or at image level (\eg, SAD, ZNCC\dots), and $\lambda_{reg}, \lambda_{emb}$ two hyper parameters that weigh the different terms of the loss function. We refer the reader to \autoref{sec:implementation} for the actual $d(x,y,),\lambda_{reg},\lambda_{emb}$ and $\phi(x,y)$ used in the experiments. 
The contribution of the three terms can be summarized as follows. $L_{adv}$ is the standard adversarial loss for the generator network that forces the synthesized images to be indistinguishable from those sampled from domain $\mathcal{B}$; $L_{reg}$ is aimed at synthesizing images that preserve the overall structure of the input ones (avoiding thereby the mode collapse issue often occurring in GAN generators);  $L_{emb}$ forces an additional adversarial behaviour against the \textit{encoder}, so as to create hard to embed samples.  

At training time, given a minibatch of $M$ different triplets of samples $(i^\mathcal{A}_p,i^\mathcal{A}_n,i^B)$, the three networks are trained jointly to minimize their average loss on the $M$ samples.

\section{Experimental Results}
\label{sec:experimental results}
\begin{figure}
	\centering	
	\includegraphics[width=0.4\textwidth]{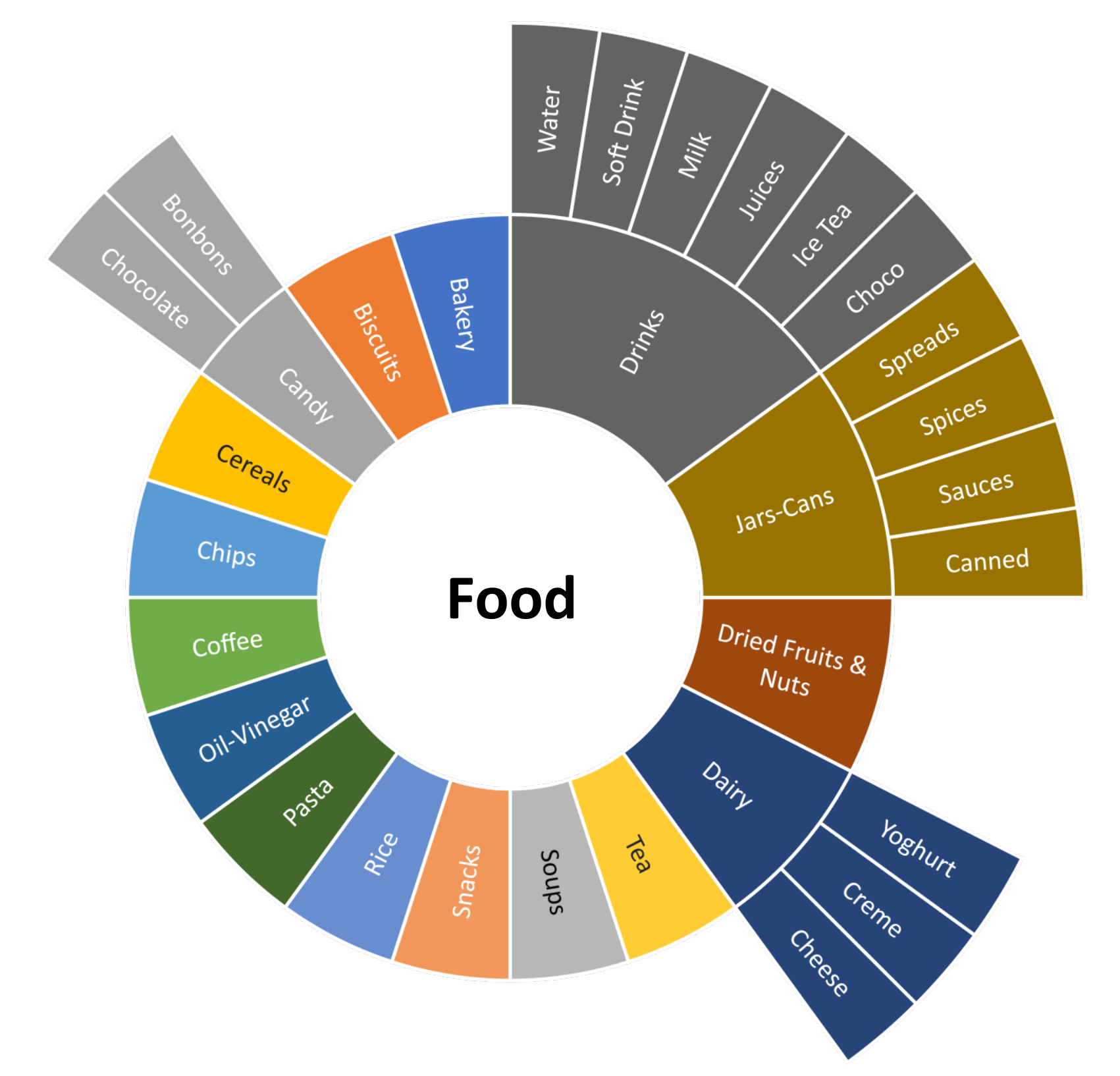}
	\caption{Visualization of the hierarchy of categories of the \textit{Grocery\_Food} dataset used as training set throughout our experiments. Each outermost category contains several different fine classes (products) not depicted for clarity.}
	\label{fig:taxonomy}
\end{figure}

\subsection{Implementation Details}
\label{sec:macro_implementation}

\paragraph{\textbf{Datasets}}
\label{sec:dataset}
To evaluate the effectiveness of \algoname{} in recognizing grocery products we rely on two products datasets comprising thousands of items: the publicly available \emph{Grocery Products} dataset (\cite{george2014recognizing}) and a standard commercial database, referred to here as \emph{Product8600}. Both datasets include more than 8500 grocery products, each described by exactly one studio-quality (\emph{reference}) image of the frontal face of the package, and feature a multi level class hierarchy in the categorization of products. As already discussed, at test time we pursue recognition from a different set of images (\emph{query}). To create this set, for \emph{Grocery Products} we automatically cropped individual items from the available shelf images according to the annotation released in \cite{tonioni2017product}, thereby obtaining  a total of 938 \emph{query} images. As for \emph{Product8600}, we cropped and annotated individual items from shelf videos that we acquired in a grocery store by a tablet camera, for a total number of 273 \emph{query} images. 
As the shelf images available in \emph{Grocery Products} concern only items belonging to the \textit{Food} macro class, which accounts for 3288 products, we consider also this smaller subset of products, which will be referred to as \emph{Grocery\_Food}, whilst \emph{Grocery\_Full} will denote all the products of the \emph{Grocery Product} dataset. We depict in \autoref{fig:taxonomy} the taxonomy of macro categories that compose the \textit{Grocery\_Food} dataset which we are going to use as training set, each categories features several fine grained classes (one for product) not depicted in figure. 
As for the samples from domain $\mathcal{B}$ needed to train the \textit{discriminator} and \textit{generator} of our architecture (see \autoref{ss:dom_inv}), we have used 547 additional images cropped from the shelf images available in \textit{Grocery Products}, picked as to have no overlap with the previously mentioned 938 \emph{query} images used at test time. 
We wish to point out how our formulation requires images from domain $\mathcal{B}$ only for the discriminator of the GAN system. Therefore, few samples without any kind of annotation are sufficient to learn the appearance of products on the shelf. Moreover we can use images from domain $\mathcal{B}$ that depict any kind of product, even items not in $\mathcal{A}$.

\paragraph{\textbf{Network architectures}}
\label{sec:network}
For the implementation of \algoname{} we have used tensorflow\footnote{\url{https://www.tensorflow.org/}} as our deep learning framework. 
For all our test we used as \textit{generator} U-Net (\cite{ronneberger2015u}) and as \textit{discriminator} PatchGAN  (\cite{isola2016pix}), the latter producing a dense grid of predictions for each input image. For the \textit{encoder} we tested different available CNN model with or without pretrained weight on the ImageNet-1000 classification task.
For the initialization of the \textit{encoder} network on the fine tuning tests we have used the weights publicly available in the tensorflow/models repository\footnote{\url{https://github.com/tensorflow/models/tree/master/research/slim}}. The three network that compose \algoname{} can easily fit in a single GPU, so training our system, once implemented, in a deep learning framework is straightforward.

\paragraph{\textbf{Descriptor Computation}}
\label{sec:descriptor}
We will show how for the grocery product recognition scenario, the best performance can be obtained using as embedding the maximum activation of convolution features (MAC \cite{tolias2015particular}). We extract these descriptors from different layers, concatenate them and finally perform L2-normalization to get a final representation laying on the unit hyper-sphere. For all our tests, we used as distance function $d(X,Y)=1-X \cdot Y$ with $X,Y \in \mathcal{D}$ (\ie, one minus the cosine similarity between the two descriptors).

\paragraph{\textbf{Training Details}}
\label{sec:implementation}
As for  $\phi$ in $L_{reg}$, we tried the pixel-wise $L1$ or $L2$ norms, the Structured Similarity Index (SSIM) (\cite{wang2004image}) and the Zero Mean Normalized Cross Correlation (ZNCC) and found out the last to work best in all our tests, in the following $\phi(x,y)=ZNCC(x,y)$. The weights of each network are trained to minimize their specific loss functions as introduced in \autoref{sec:metodo}. We use Adam (\cite{adam}) as optimizer with different learning rates for the different tests. Concerning data preprocessing, we use as input colour images with fixed size of $256 \times 256$ and intensities rescaled between $[-1,1]$.  To obtain the input dimension the original images are rescaled to the target resolution preserving the aspect ratio and filling the extra pixels with 0s. The only additional data augmentation is a preliminary random crop with size at least $80\%$ of the original image to attain the input of the \textit{generator} network.

\paragraph{\textbf{Evaluation Protocol}}
\label{sec:protocol}
To test our embedding network we encode all the \emph{reference} images of the considered dataset to create a reference database, then compute the same encoding for the \emph{query} images. For each query vector we perform similarity search against all the reference vectors and retain the K most similar database entries; if the reference image for the product depicted in the query does belong to this set of nearest neighbours we consider recognition to be successful. As a measure of effectiveness of the embedding, we report the accuracy (number of successful recognitions over number of queries) for different K values.

Based on these premises, we train once and for all our architecture using only the reference images belonging to \emph{Grocery\_Food}, (\ie, one reference image for each of 3288 different products organized in a multi level hierarchy of products categories). We then use the trained embedding model to address three different test scenarios:    
\begin{enumerate}[(a)]
	\item \emph{Grocery\_Food}: we recognize the 938 \emph{query} images from \emph{Grocery Products} based on the 3288 reference images from \emph{Grocery\_Food}. Thus, all the \textit{reference} images were deployed at training time. 
	\item \emph{Grocery\_Full}: we recognize the 938 \emph{query} images from \emph{Grocery Products} based on the 8403 reference images from \emph{Grocery\_Full}. Thus, only 40\% of the \textit{reference} images were deployed at training time. 
	\item \emph{Product8600}: we recognize the 273 \emph{query} images cropped from our videos based on the 8597 reference images from \emph{Product8600}. Thus, none of the \textit{reference} images was deployed at training time. 
\end{enumerate}

Among the three, (b) is the most likely to happen in practical settings as product appearance changes frequently overtime and it is infeasible to constantly retrain the embedding network, although perhaps a portion of the reference images dealing with the items actually on sale in the store had been used at training time

At test time the \textit{encoder} network is used as a stand alone global image descriptor. Our implementation is quite efficient and can easily encode, on GPU, more than 200 image per second taking into account also the time needed to load the images from disk and rescale them to the $256 \times 256$ input size.
Finally, given that usually our \textit{reference} database only provides a single image per product and the descriptor dimension is relatively low (between 256 and 1024 floats across different tests), we perform the K-NN similarity search extensively without any kind of approximation. The biggest descriptor database considered in our tests is the one obtained from \textit{Product8600} using a 1024 float dimensional embedding vector (\ie, the \textit{reference} database is a matrix of float with 8600 row and 1024 column). Even on this kind of database given a query descriptor the whole similarity search can be solved in a tenth of second using brute force search, nevertheless the search could be speeded up using KD-Tree or approximate search technique.

\subsection{Ablation Study}
\label{ss:ablation}

\begin{table*}
	\caption{\label{tab:ablation} Ablation study for \algoname{}. Recognition accuracy for 1-NN and 5-NN similarity search in the three considered scenarios. Best results highlighted in bold.}
	\centering
	\begin{tabular}{clcccccc}
		\toprule
		&& \multicolumn{2}{c}{(a) \textit{Grocery\_Food}}&\multicolumn{2}{c}{(b) \textit{Grocery\_Full}}&\multicolumn{2}{c}{(c) \textit{Product8600}}\\
		\midrule
		&\textit{Training loss}&K=1&K=5&K=1&K=5&K=1&K=5\\
		\midrule
		(1) & triplet&0.301&0.430&0.277&0.390&0.351&0.490\\
		(2) & hierarchy&0.325&0.491&0.302&0.433&0.355&0.553\\
		(3) & triplet+GAN&0.454&0.626&0.418&0.586&0.512&0.706\\
		(4) & hierarchy+GAN&0.479&0.660&0.455&0.621&0.538&0.699\\
		(5) & triplet+GAN+adv&0.470&0.648&0.431&0.595&0.548&0.717\\
		(6) & hierarchy+GAN+adv (\textbf{\algoname{}})&\textbf{0.481}&\textbf{0.688}&\textbf{0.463}&\textbf{0.642}&\textbf{0.553}&\textbf{0.732}\\
		\bottomrule
		
	\end{tabular}	
\end{table*}

To understand the impact on performance of the different novel components proposed in our architecture, we carry out a model ablation study using as \textit{encoder} PatchGAN (\cite{isola2016pix}) with MAC features (\cite{tolias2015particular}) extracted from the last convolutional layer before the output. 
We use this randomly initialized small network to better highlight the gains provided by the different kind of proposed losses. We will show how to obtain the best performance we rely on a larger pre-initialized network. 
We train this architecture on the \textit{reference} images of \textit{Grocery\_Food} according to six different training losses and report the accuracy dealing with the three test scenarios presented in \autoref{sec:experimental results} in \autoref{tab:ablation}. In particular, with reference to the first column, \textit{triplet} denotes training by triplet loss with fixed margin ($\alpha=0.3$ obtained by cross validation); \textit{hierarchy} denotes training by our triplet loss with variable margin introduced in \autoref{ss:hi_emb} ($\alpha_{min}=0.1,\alpha_{max}=0.5$); entries with \textit{+GAN} denote deploying the image translation GAN to generate the anchor image $i^\mathcal{B}_a$ (\autoref{eq:gen_loss}: $\lambda_{reg}=1,\lambda_{emb}=0$); finally, entries with \textit{+GAN+adv} concerns introducing also the adversarial term in the loss of the GAN generator (\autoref{eq:gen_loss}: $\lambda_{reg}=1,\lambda_{emb}=0.1$). For all the models that do not use GANs, we rely on standard data augmentation techniques (\eg, crop, gaussian blur, color transformation\dots) to obtain the anchor image given the positive one. In all the tests the networks are randomly initialized and trained for the same number of steps and identical learning rates.

The results reported in \autoref{tab:ablation} show how each individual novel component proposed in our  \algoname{} architecture provides a significant performance improvement with respect to the standard triplet ranking loss. Indeed, by comparing rows (2),(4) and (6) to (1),(3) and (5), respectively, it can be observed that modifying the fixed margin of the standard triplet loss into our proposed hierarchically adaptive margin can improve accuracy with all models and in all scenarios, with a much larger gain in (c)(\ie, completely unseen \textit{reference} images). This proves that embedding a hierarchy into the descriptor space is an effective strategy to help learning an embedding amenable to generalize to unseen data.
The main improvements are clearly achieved by methods featuring a GAN network to pursue domain adaptation by generating training samples similar to the images coming from the test domain. Indeed, comparison of  (3) and (4) to (1) and (2), respectively, highlights how performance nearly double across all models and scenarios, testifying that the domain shift between test and train data and the lack of multiple training samples are indeed the key issues in this task that the proposed image-to-image translation GAN can help to address very effectively. 
Finally, comparing  (5) and (6) to (3) and (4), respectively, vouches that training the \textit{generator} to produce  anchor images not only realistic but also hard to embed turns out always beneficial to performance. Indeed, the adversarial game played by the \textit{generator} and \textit{encoder} may be thought of as an on-line and adaptive hard-mining capable of dynamically synthesize hard to embed samples that help training a more robust embedding. Performance of our overall  \algoname{} architecture are reported in row (6) and show a dramatic improvement with respect to the standard triplet loss, row (1).  

\subsection{Product Recognition}
\label{ss:fine}
\paragraph{\textbf{Model and Descriptor Selection}}
To ameliorate product recognition performance we can rely on larger networks pre-trained on the ImageNet classification benchmark. To chose the best CNN model as our \textit{encoder} network we downloaded the public available weights of different models trained on ImageNet-1000 classification and test them as general purpose off-the-shelf feature extractors on our datasets without any kind of fine tuning. We considered three different popular CNN models (VGG\_16 \cite{Simonyan14c}, resnet\_[50/101/152] \cite{he2016deep} and inception\_v4 \cite{szegedy2017inception}) and compute three kind of different descriptors from activations extracted at various layers:
\begin{itemize}
	\item \textbf{Direct}: directly use the vectorized activation of a given layer. The dimension of the descriptor is the number of elements in the feature maps for that layer. 
	\item \textbf{AVG}: perform average pooling on the feature map with a kernel with width and height equals those of the map. Therefore, we use as descriptor the average activation of each convolutional filter for a given layer. The dimension of the descriptor is the number of convolutional filters in the selected layer. 
	\item \textbf{MAC [\cite{tolias2015particular}]}: perform max pooling on the feature map with a kernel with width and height equals those of the map. Therefore, we use as descriptor the maximum activation of each convolutional filter for a given layer. The dimension of the descriptor is the number of convolutional filters in the selected layer.
\end{itemize}

We applied the three different descriptors above at different layers of the three networks and report in \autoref{tab:modelStudy} the 1-NN accuracy using the test protocol described in \autoref{ss:ablation}. For all our tests the descriptors were L2 normalized to unit norm and the similarity search is performed using cosine similarity. In \autoref{tab:modelStudy} we report only some of the best performing layers for each network and additional tests where the descriptors are obtained by concatenation of representations extracted at different depths in the CNN (\eg, \textit{conv4\_3+conv5\_3} is the concatenation of representations extracted at layers conv4\_3 and conv5\_3) with L2 normalization performed after concatenation.

\begin{table*}
	\centering
	\caption{\label{tab:modelStudy} 1-NN accuracy for different general purpose descriptors obtained from layers of network pre-trained on the ImageNet-1000 classification dataset without any kind of additional fine-tuning. Best results are higlighted in bold.}
	\scalebox{1}{
		\begin{tabular}{ccc|ccc}
			\toprule
			\textbf{Network}&\textbf{Layer}&\textbf{Descriptor}&\textbf{Grocery\_Food}&\textbf{Grocery\_Full}&\textbf{Product8600}\\
			\midrule
			\multirow{8}{*}{\emph{VGG\_16}}&\multirow{2}{*}{conv4\_3}&MAC&0.789&0.785&0.717\\
			&&AVG&0.515&0.510&0.538\\
			&\multirow{2}{*}{conv5\_3}&MAC&0.724&0.720&0.611\\
			&&AVG&0.406&0.398&0.395\\
			&\multirow{2}{*}{conv4\_3+conv5\_3}&MAC&\textbf{0.792}&\textbf{0.787}&\textbf{0.725}\\
			&&AVG&0.501&0.493&0.523\\
			&fc6&Direct&0.560&0.549&0.549\\
			&fc7&Direct&0.444&0.433&0.432\\
			\midrule
			\multirow{6}{*}{\emph{inception\_v4}}&\multirow{2}{*}{Mixed\_7a}&MAC&0.610&0.603&0.509\\
			&&AVG&0.673&0.670&0.560\\
			&\multirow{2}{*}{Mixed\_7b}&MAC&0.652&0.641&0.512\\
			&&AVG&0.675&0.668&0.5091\\
			&\multirow{2}{*}{Mixed\_7a+Mixed\_7b}&MAC&0.655&0.646&0.534\\
			&&AVG&0.690&0.685&0.542\\
			\midrule
			\multirow{6}{*}{\emph{resnet\_50}}&\multirow{2}{*}{Block3}&MAC&0.731&0.729&0.703\\
			&&AVG&0.441&0.433&0.432\\
			&\multirow{2}{*}{Block4}&MAC&0.654&0.646&0.509\\
			&&AVG&0.571&0.558&0.465\\
			&\multirow{2}{*}{Block3+Block4}&MAC&0.723&0.720&0.644\\
			&&AVG&0.547&0.538&0.545\\
			\midrule
			\multirow{6}{*}{\emph{resnet\_101}}&\multirow{2}{*}{Block3}&MAC&0.737&0.735&0.695\\
			&&AVG&0.389&0.388&0.432\\
			&\multirow{2}{*}{Block4}&MAC&0.636&0.662&0.490\\
			&&AVG&0.570&0.556&0.417\\
			&\multirow{2}{*}{Block3+Block4}&MAC&0.714&0.708&0.626\\
			&&AVG&0.535&0.524&0.520\\
			\midrule
			\multirow{6}{*}{\emph{resnet\_152}}&\multirow{2}{*}{Block3}&MAC&0.708&0.703&0.655\\
			&&AVG&0.345&0.337&0.446\\
			&\multirow{2}{*}{Block4}&MAC&0.571&0.561&0.435\\
			&&AVG&0.571&0.561&0.435\\
			&\multirow{2}{*}{Block3+Block4}&MAC&0.678&0.671&0.542\\
			&&AVG&0.506&0.500&0.504\\
			\bottomrule
		\end{tabular}
	}
\end{table*}

\begin{table*}
	\caption{\label{tab:ablation_vgg} Ablation study for \algoname{} on a VGG-16 network pretrained on ImageNet-1000. Recognition accuracy for 1-NN and 5-NN similarity search in the three considered scenarios. Best results highlighted in bold.}
	\centering
	\begin{tabular}{clcccccc}
		\toprule
		&& \multicolumn{2}{c}{(a) \textit{Grocery\_Food}}&\multicolumn{2}{c}{(b) \textit{Grocery\_Full}}&\multicolumn{2}{c}{(c) \textit{Product8600}}\\
		\midrule
		&\textit{Training loss}&K=1&K=5&K=1&K=5&K=1&K=5\\
		\midrule
		(1) & triplet&0.799&0.922&0.775&0.894&0.765&0.915\\
		(2) & hierarchy&0.812&0.933&0.816&0.926&0.805&0.952\\
		(3) & triplet+GAN&0.829&0.941&0.821&0.937&0.816&0.945\\
		(4) & hierarchy+GAN&0.832&0.943&0.826&0.933&0.819&0.952\\
		(5) & triplet+GAN+adv&0.833&\textbf{0.948}&0.821&0.937&0.816&0.945\\
		(6) & hierarchy+GAN+adv (\textbf{\algoname{}})&\textbf{0.853}&\textbf{0.948}&\textbf{0.842}&\textbf{0.942}&\textbf{0.827}&\textbf{0.959}\\
		\bottomrule
		
	\end{tabular}	
\end{table*}

Looking at the results in \autoref{tab:modelStudy} we can observe how, in our settings, newer and more powerful CNN, like inception\_v4 or resnet\_152, fail to achieve the same instance-level distinctiveness of  VGG\_16. 
We conjecture that deeper architectures, trained on Imagenet, tend to create more abstract representations that may not provide out-of-the-box features distinctive enough to tell apart many items looking almost identical as required by our problem. Some evidence to support this conjecture may be found in \autoref{tab:modelStudy} due to deeper layers providing tipically inferior performace when compared to shallower ones (e.g., VGG\_16: conv5\_3 vs conv4\_3, resnet\_50,resnet\_101,resnet\_152: Block4 vs Block3)
Concerning the type of descriptor to use, from \autoref{tab:modelStudy} it seems quite clear that MAC descriptor is the best choice for grocery recognition with respect to AVG or direct activations of fully connected layers. 

Given these results, we selected for our fine tuning tests the VGG\_16 network with MAC descriptor computed on the concatenation of conv4\_3 and conv5\_3 layers, and train the overall architecture according to our losses. As the \textit{encoder} is already pre-trained, we perform 5000 iterations of pre-training for the \textit{generator} and \textit{discriminator} (\autoref{eq:gen_loss}: $\lambda_{reg}=1,\lambda_{emb}=0$) before training jointly the whole \algoname{} architecture. The 
chosen hyper-parameters obtained by cross validation for the training process are as follows. Learning rates $10^{-5}$,$10^{-5}$ and $10^{-6}$ for \textit{generator}, \textit{discriminator} and \textit{encoder}, respectively; $\lambda_{emb}=0.1$, $\lambda_{reg}=1$, $\alpha_{min}=0.05$ and $\alpha_{max}=0.5$.

Before comparing our proposal to other embedding losses, it is interesting to verify whether the improvements provided by the different components (\autoref{ss:ablation}) are valid even when relying on the VGG-16 network pretrained on Imagenet. 
Purposely, we carry out the same ablation study as in \autoref{tab:ablation} and report the results in \autoref{tab:ablation_vgg}.
Indeed, the ranking of performances among the different training modalities turns out coherent, although, as expected, the margins are smaller due to the higher performance provided by the baseline. 


\paragraph{\textbf{Comparison with other embedding losses}} We compare our architecture to the already mentioned concatenation of MAC descriptors without fine tuning (\textit{MAC}) and to our implementation of different embedding learning methods: \cite{wang2014learning} fine tuning using the classic triplet ranking loss (\textit{Triplet}); \cite{hadsell2006dimensionality} fine tuning using Siamese networks and the contrastive loss (\textit{Siamese}); Matching networks \cite{vinyals2016matching} without the \textit{full context embedding} which does not scale to thousands of classes (\textit{MatchNet}); \cite{zhang2016embedding}  tuplet loss to embed label structure (\textit{Structured}); and \cite{zhang2017learning}  triplet loss regularized by a spread out term (\textit{Spread}). Similarly to \algoname{}, all methods are trained on the \textit{reference} images of \textit{Grocery\_Food} starting from the very same VGG16 pre-trained on ImageNet-1000 and using the same concatenation of MAC descriptors as embedding function.

\begin{table*}
	\caption{\label{tab:results} Recognition accuracy for 1-NN and 5-NN similarity search in the three considered scenarios. Best results highlighted in bold, differences between DIHE and best performing competitor reported in the last line.}
	\centering
	\begin{tabular}{ccccccc}
		\toprule
		& \multicolumn{2}{c}{(a) \textit{Grocery\_Food}}&\multicolumn{2}{c}{(b) \textit{Grocery\_Full}}&\multicolumn{2}{c}{(c) \textit{Product8600}}\\
		\midrule
		\textit{Descriptor}&K=1&K=5&K=1&K=5&K=1&K=5\\
		\midrule
		MAC&0.792&0.917&0.787&0.9093&0.725&0.908\\
		\midrule
		Triplet [\cite{wang2014learning}]&0.799&0.922&0.775&0.894&0.765&0.915\\
		Spread [\cite{zhang2017learning}]&0.784&0.916&0.764&0.893&0.758&0.923\\	
		Structured [\cite{zhang2016embedding}]&0.809&0.931&0.804&0.926&0.750&0.912\\
		Siamese [\cite{hadsell2006dimensionality}]&0.810&0.931&0.805&0.928&0.733&0.926\\
		MatchNet [\cite{vinyals2016matching}]&0.834&0.939&0.810&0.929&0.820&0.948\\
		\midrule
		\multirow{2}{25 pt}{\algoname{}}&\textbf{0.853} &\textbf{0.948}&\textbf{0.842}&\textbf{0.942}&\textbf{0.827}&\textbf{0.959}\\
		&+0.02&+0.01&+0.03&+0.02&+0.007&+0.01\\
		\bottomrule
		
	\end{tabular}
\end{table*}

\begin{figure*}
	\begin{tabular}{ccc}
		\includegraphics[width=0.31\textwidth]{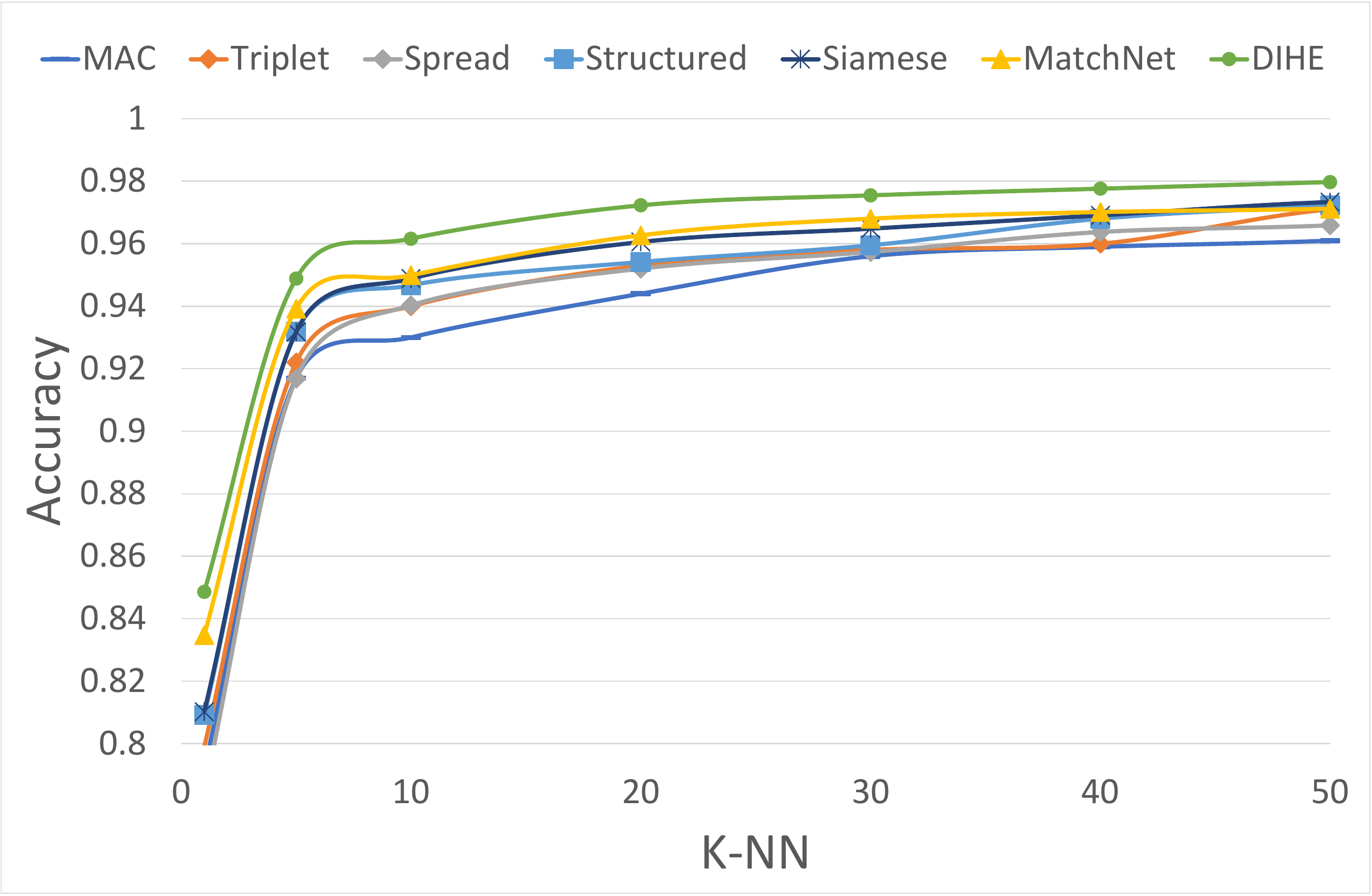}&
		\includegraphics[width=0.31\textwidth]{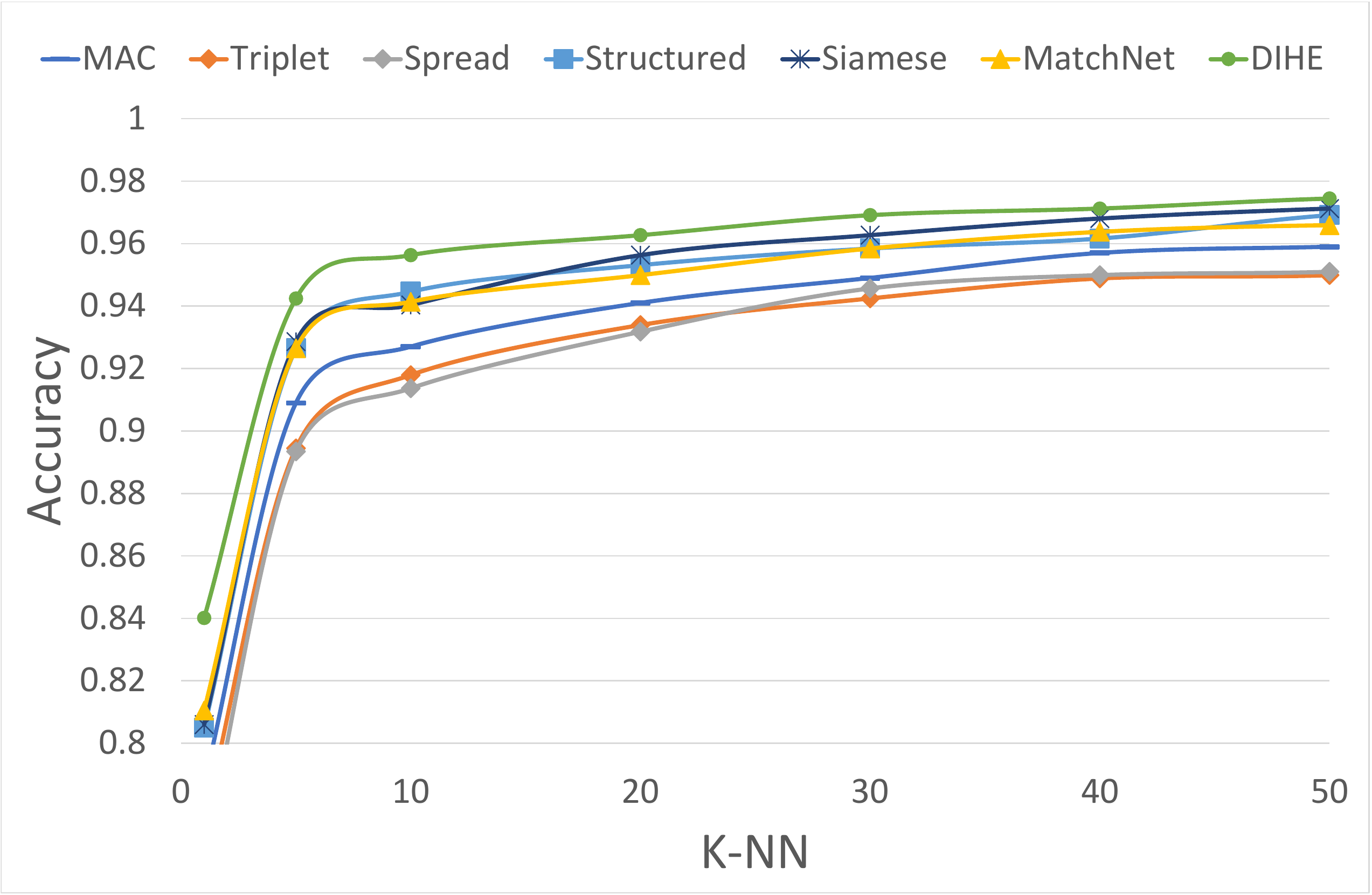}&
		\includegraphics[width=0.31\textwidth]{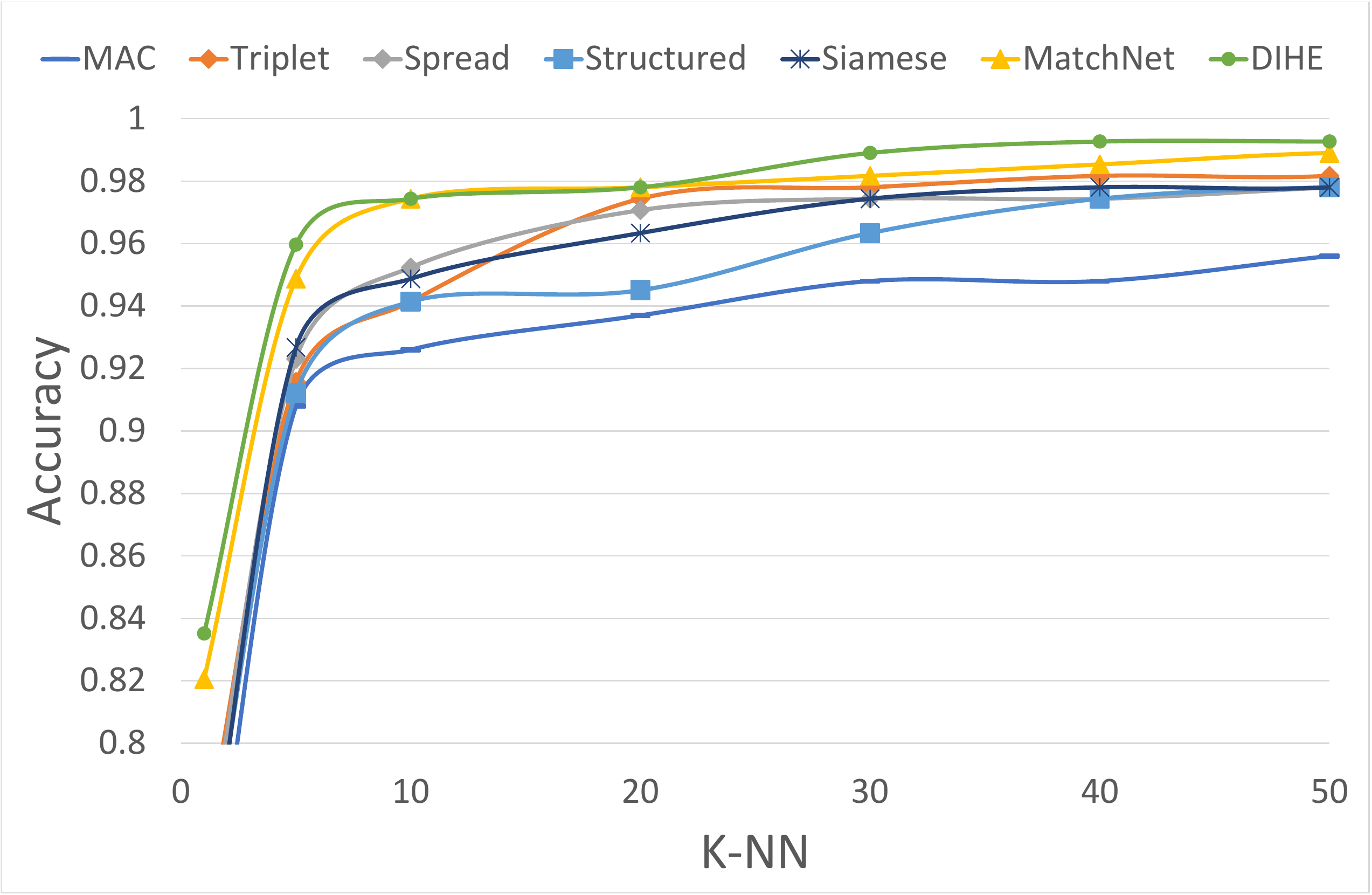}\\
		(a) \textbf{Grocery\_Food} & (b) \textbf{Grocery\_Full} & (c) \textbf{Product8600}\\
	\end{tabular}
	\caption{Accuracy with increasing K in the three scenarios.}
	\label{fig:accuracy_k}
\end{figure*}

We use the same test scenarios presented in \autoref{ss:ablation} and report the result in \autoref{tab:results}. As already shown in our model study, MAC activations have strong absolute performance without any need of fine tuning with accuracy at K=1 ranging from 0.72 in the worst case to 0.79 in the best. Starting from such a strong baseline the standard \textit{Triplet} loss is able to only slightly increase performance in scenario (a) and (c) while being slightly penalized in (b), which testifies how in the cross domain and low shot regime is quite hard to properly fine tune an embedding network. The additional regularization term introduced by \textit{Spread} does not seems to help with a slightly decrease in performance across all scenario compared with the standard \textit{Triplet}. \textit{Structured} is the first method to consistently improve performance across all scenario, vouching the importance of deploying label structure in the embedding space for this type of recognition task. \textit{Siamese} based on a simple contrastive loss is able to obtain performance comparable to \textit{Structured} on scenario (a) and (b) while performing slightly worse on the generalization to dataset (c). \textit{MatchNet} is definitely the best competing method for embedding learning in a few shot regime as testified by the good improvement obtained with respect to the initial MAC descriptors. Nevertheless, \algoname{} thanks to the combined use of GAN and hierarchical information is still able to improve the performance, obtaining recognition accuracies for K=1  consistently over 82\% across all test.   It is worth pointing out that the largest improvement, with respect to the inital MAC results, is obtained on the completely unseen products of the \textit{Product8600}, which vouches for mixing GAN-based domain adaptation and hierarchical embedding to learn an embedding that can generalize very well to unseen items. 

In \autoref{fig:accuracy_k} we report the accuracy of the methods while increasing K.  In \autoref{fig:accuracy_k}(a-b) almost all methods converge to more than 95\% accuracy for $K>30$. However, \algoname{} can provide substantially better results for lower values of K. In \autoref{fig:accuracy_k}(c) \algoname{} still outperforms the competitors with performance almost equal to those achieved by Matching Networks but showing higher margin against the competitor trained with variant of \textit{triplet ranking loss}.

\subsection{Beyond product recognition}

\begin{table}
	\caption{\label{tab:amazon_fine_res} Recognition accuracy for 1-NN and 5-NN similarity search on two subset of the \textit{Office31} using as \textit{reference} images the \textit{Amazon} subset. Best resulted highlighted in bold, differences between DIHE and best performing competitor reported in the last line.}
	\centering
	\scalebox{0.95}{
	\begin{tabular}{ccccc}
		\toprule
		& \multicolumn{2}{c}{(a) \textit{Webcam}}&\multicolumn{2}{c}{(b) \textit{DSLR}}\\
		\midrule
		\textit{Descriptor}&K=1&K=5&K=1&K=5\\
		\midrule
		FC6&0.470&\textbf{0.698}&0.489&0.712\\
		MAC&0.382&0.640&0.393&0.660\\
		\midrule
		Triplet [\cite{wang2014learning}]&0.596&0.675&0.628&0.710\\
		Spread [\cite{zhang2017learning}]&0.591&0.660&0.590&0.670\\
		Siamese [\cite{hadsell2006dimensionality}]&0.469&0.547&0.576&0.630\\
		MatchNet [\cite{vinyals2016matching}]&0.522&0.579&0.566&0.618\\
		\midrule
		\multirow{2}{25 pt.}{\algoname{}}&\textbf{0.628}&0.691&\textbf{0.662}&\textbf{0.742}\\
		&+0.03&-0.007&+0.03&+0.03\\
		\bottomrule		
	\end{tabular}
	}
\end{table}

To investigate on the generality of our proposal as an improvement over established embedding learning methods, we perform additional tests on the \textit{Office31} benchmark dataset for domain adaptation (\cite{saenko2010adapting}). This dataset  consists of 4652 images dealing with 31 classes of office objects acquired in three different domains, namely \textit{Amazon}, concerning ideal images downloaded from the web, \textit{Webcam}, consisting of real images acquired by cheap cameras, and \textit{DSLR}, featuring real images acquired by an high quality camera. Akin to the setup of Grocery Recognition experiments,  we use \textit{Amazon} as the \textit{reference} set, thus deploying these images to train the \textit{encoder}, while images from \textit{Webcam} and \textit{DSLR} are used both as training samples for the GAN \textit{discriminator} in \algoname{}  and as two separate \textit{query} sets for the tests. This scenario is  compliant to the \textit{full protocol setting} described by \cite{gong2013connecting}: train on the entire labeled source and unlabeled target data and test on annotated target samples. Unfortunately, as the \textit{Office31} dataset does not provide a taxonomy of classes, in these additional experiments  \algoname{} can not leverage on the hierarchical loss to improve the learned representation. However, unlike the Grocery Recognition scenario, all methods can be trained here by more than one sample per class. 

Based on the above described experimental setup, we train different \textit{encoders} starting, once again, from a VGG16 network pre-trained on the ImageNet-1000 dataset. We report the accuracy for 1-NN and 5-NN similarity search in \autoref{tab:amazon_fine_res}. To asses the performance of \algoname{} we compare it once again against the methods considered in \autoref{ss:fine}, with the exception of \textit{Structured} due to \textit{Office31} lacking a class taxonomy. The first two rows of \autoref{tab:amazon_fine_res} show that, differently from the the Grocery Recognition setup, between activations extracted from the pretrained VGG16 network, FC6 outperforms  MAC descriptors (computed as in \autoref{ss:fine}). 
Coherently with the findings of \autoref{tab:modelStudy}, we believe this difference to be due to the office task concerning the recognition of category of objects rather than instances. Accordingly, to obtain the \textit{encoder} network for these tests we substitute the original FC6 layer with a smaller fully connected layer consisting of 512 neurons with randomly initialized weights and then perform fine tuning on the \textit{Amazon} images.

In this different experimental settings wherein more training samples per class are available, both \textit{Siamese}, \textit{Spread Out} and \textit{Matchnet} are outperformed by the plain \textit{Triplet} ranking loss of  \cite{wang2014learning}, which turns out the best performing established method achieving a 1-NN accuracy of 59\% and 62\% for test datasets \textit{Webcam}  and \textit{DSLR}, respectively. Yet, thanks to the introduction of the \textit{Generator} in the training loop, \algoname{} can yield a significant performance improvement reaching a 1-NN accuracy of 62\% (+3\%) and 66\% (+4\%) for \textit{Webcam}  and \textit{DSLR}, respectively, with best or comparable 5-NN accuracy when compared with competing methods using descriptors with the same dimension. Moreover, our proposal can outperform the descriptor extracted by FC6 that is twice larger on 3 experiments out of 4, obtaining comparable performance on the fourth. 

Although performance on \textit{Office31} turns out well below the state-of-the-art attainable by image recognition methods based on classifiers, amenable to handle a fixed and possible small number of classes, we argue that the experiments reported in this Section further highlight the advantages provided by our proposed \algoname{} architecture with respect to common feature learning approaches. 

\subsection{Qualitative Results}
\label{sec:qualitativi}
\begin{figure*}
	\centering
	\begin{tabular}{c|ccccc}
		\toprule
		\multicolumn{6}{c}{\textbf{Grocery\_Food}}\\
		\midrule
		\textbf{Query} & \textbf{1-NN} & \textbf{2-NN} & \textbf{3-NN} & \textbf{4-NN} & \textbf{5-NN} \\
		\includegraphics[width=0.14\textwidth]{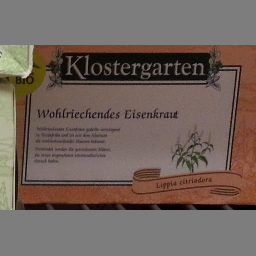} & \cellcolor{red!16!green!68!blue}\includegraphics[width=0.14\textwidth]{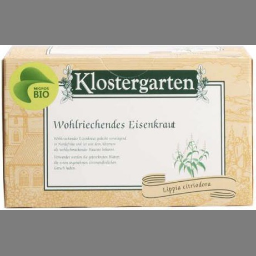} & \includegraphics[width=0.14\textwidth]{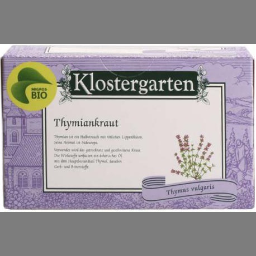} & \includegraphics[width=0.14\textwidth]{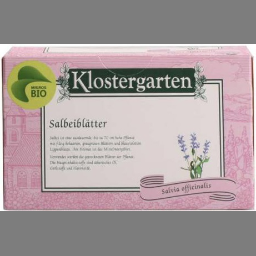}	& \includegraphics[width=0.14\textwidth]{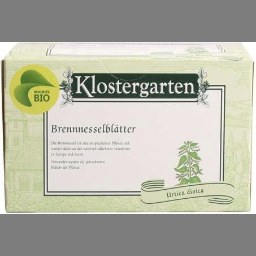} & \includegraphics[width=0.14\textwidth]{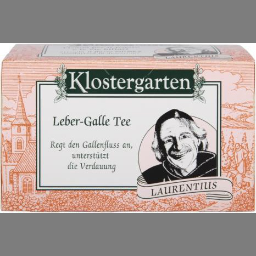} \\
		\includegraphics[width=0.14\textwidth]{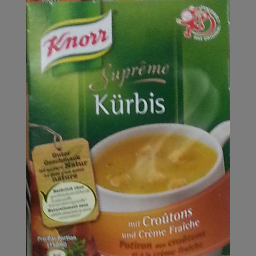} & \cellcolor{red!16!green!68!blue}\includegraphics[width=0.14\textwidth]{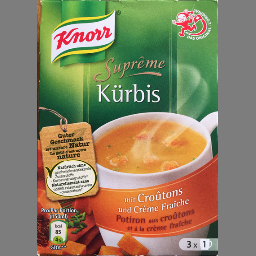} & \includegraphics[width=0.14\textwidth]{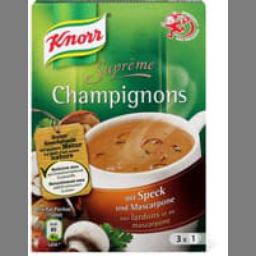} & \includegraphics[width=0.14\textwidth]{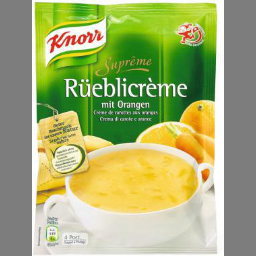}	& \includegraphics[width=0.14\textwidth]{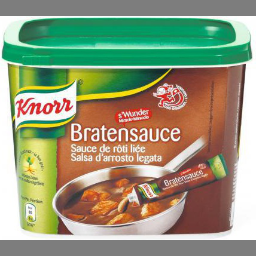} & \includegraphics[width=0.14\textwidth]{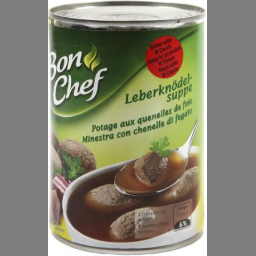} \\
		\includegraphics[width=0.14\textwidth]{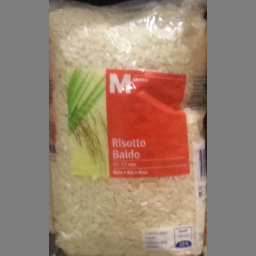} & \cellcolor{red!16!green!68!blue}\includegraphics[width=0.14\textwidth]{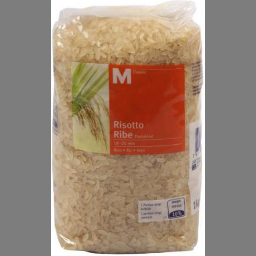} & \includegraphics[width=0.14\textwidth]{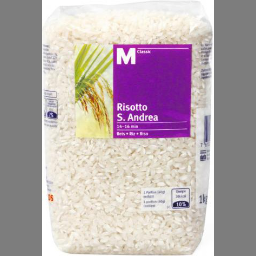} & \includegraphics[width=0.14\textwidth]{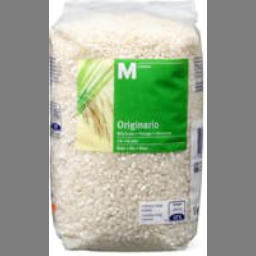}	& \includegraphics[width=0.14\textwidth]{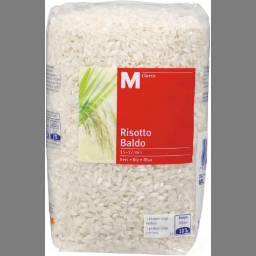} & \includegraphics[width=0.14\textwidth]{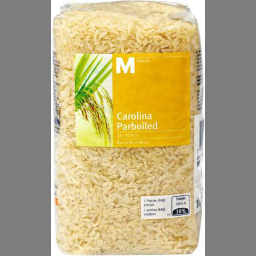} \\
	\end{tabular}
	\begin{tabular}{c|ccccc}
		\toprule
		\multicolumn{6}{c}{\textbf{Product8600}}\\
		\midrule
		\textbf{Query} & \textbf{1-NN} & \textbf{2-NN} & \textbf{3-NN} & \textbf{4-NN} & \textbf{5-NN} \\
		\includegraphics[width=0.14\textwidth]{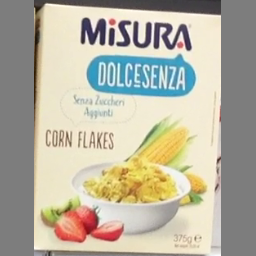} & \cellcolor{red!16!green!68!blue}\includegraphics[width=0.14\textwidth]{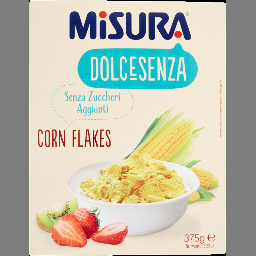} & \includegraphics[width=0.14\textwidth]{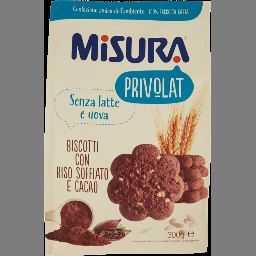} & \includegraphics[width=0.14\textwidth]{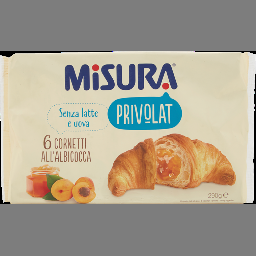}	& \includegraphics[width=0.14\textwidth]{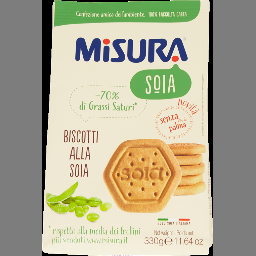} & \includegraphics[width=0.14\textwidth]{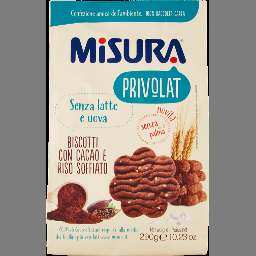} \\
		\includegraphics[width=0.14\textwidth]{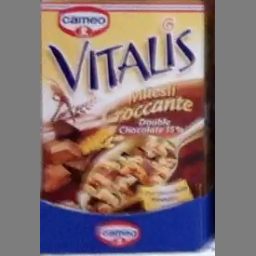} & \cellcolor{red!16!green!68!blue}\includegraphics[width=0.14\textwidth]{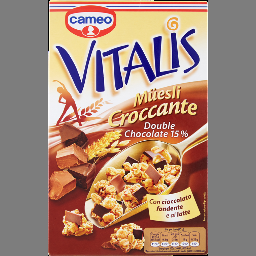} & \includegraphics[width=0.14\textwidth]{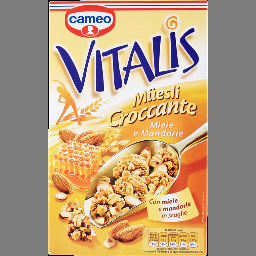} & \includegraphics[width=0.14\textwidth]{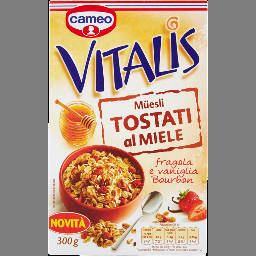}	& \includegraphics[width=0.14\textwidth]{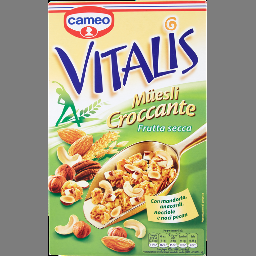} & \includegraphics[width=0.14\textwidth]{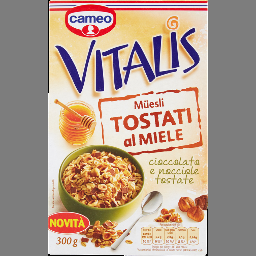} \\
		\includegraphics[width=0.14\textwidth]{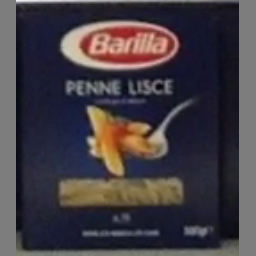} & \cellcolor{red!16!green!68!blue}\includegraphics[width=0.14\textwidth]{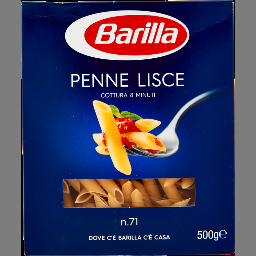} & \includegraphics[width=0.14\textwidth]{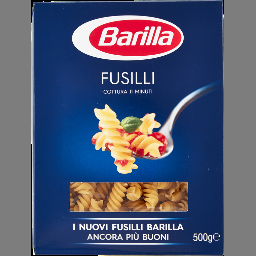} & \includegraphics[width=0.14\textwidth]{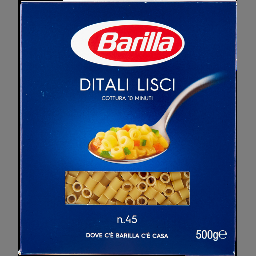}	& \includegraphics[width=0.14\textwidth]{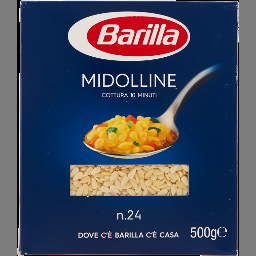} & \includegraphics[width=0.14\textwidth]{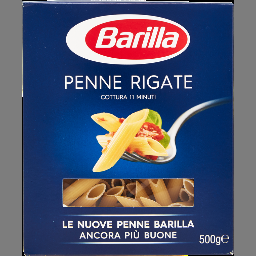} \\
	\end{tabular}
	\caption{\label{fig:qualitativi2} Qualitative results for K-NN similarity search  by  \algoname{} on \textit{Grocery\_Food} and \textit{Product8600}. Correct results highlighted in green. }
\end{figure*}

\begin{figure*}
	\centering
	\begin{tabular}{c|ccccc}
		\toprule
		\multicolumn{6}{c}{\textbf{DSLR}}\\
		\midrule
		\textbf{Query} & \textbf{1-NN} & \textbf{2-NN} & \textbf{3-NN} & \textbf{4-NN} & \textbf{5-NN} \\
		\includegraphics[width=0.14\textwidth]{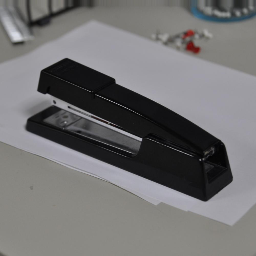} & \includegraphics[width=0.14\textwidth]{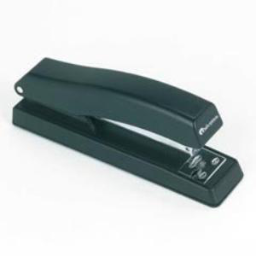} & \includegraphics[width=0.14\textwidth]{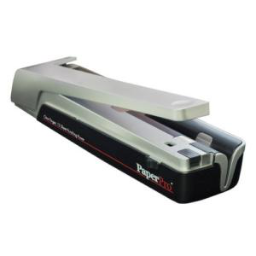} & \includegraphics[width=0.14\textwidth]{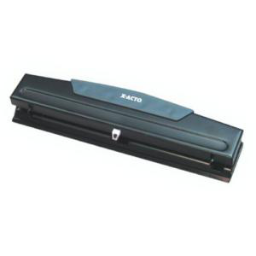}	& \includegraphics[width=0.14\textwidth]{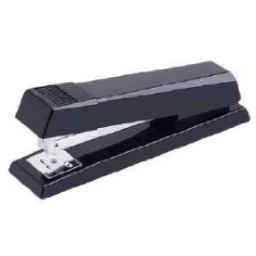} & \includegraphics[width=0.14\textwidth]{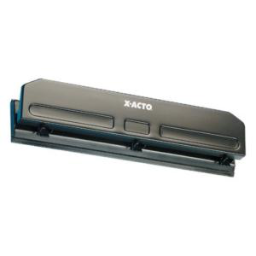} \\
		\includegraphics[width=0.14\textwidth]{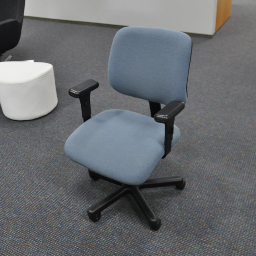} & \includegraphics[width=0.14\textwidth]{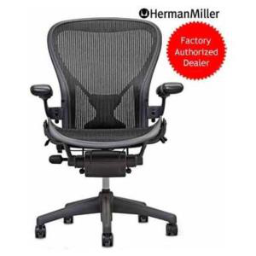} & \includegraphics[width=0.14\textwidth]{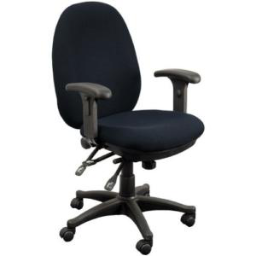} & \includegraphics[width=0.14\textwidth]{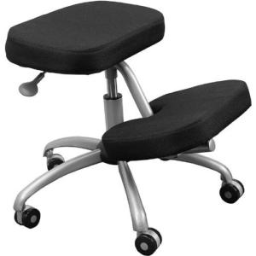}	& \includegraphics[width=0.14\textwidth]{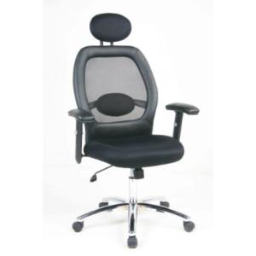} & \includegraphics[width=0.14\textwidth]{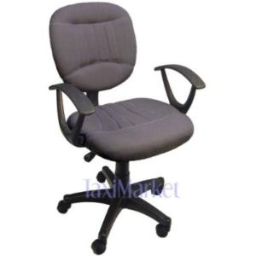} \\
		\includegraphics[width=0.14\textwidth]{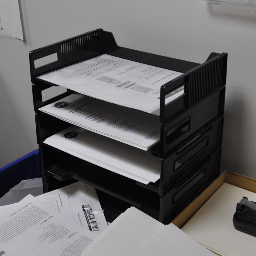} & \includegraphics[width=0.14\textwidth]{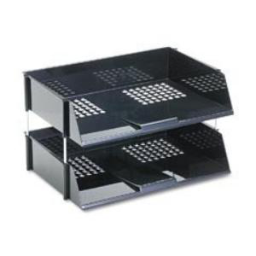} & \includegraphics[width=0.14\textwidth]{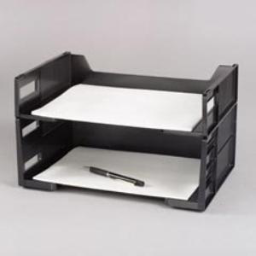} & \includegraphics[width=0.14\textwidth]{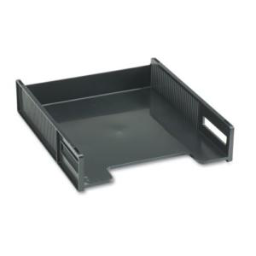}	& \includegraphics[width=0.14\textwidth]{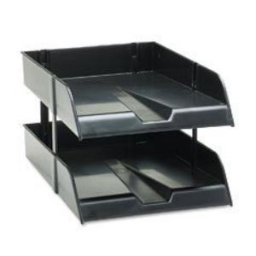} & \includegraphics[width=0.14\textwidth]{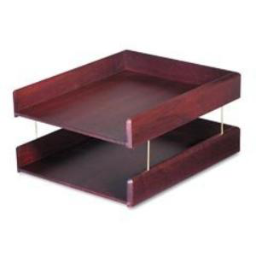} \\
	\end{tabular}
	\begin{tabular}{c|ccccc}
		\toprule
		\multicolumn{6}{c}{\textbf{Webcam}}\\
		\midrule
		\textbf{Query} & \textbf{1-NN} & \textbf{2-NN} & \textbf{3-NN} & \textbf{4-NN} & \textbf{5-NN} \\
		\includegraphics[width=0.14\textwidth]{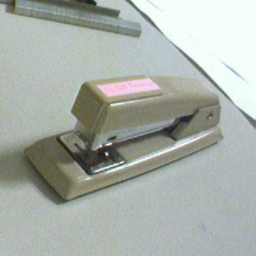} & \includegraphics[width=0.14\textwidth]{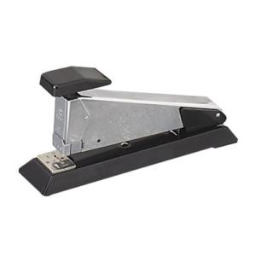} & \includegraphics[width=0.14\textwidth]{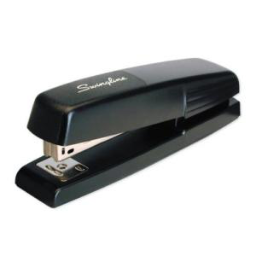} & \includegraphics[width=0.14\textwidth]{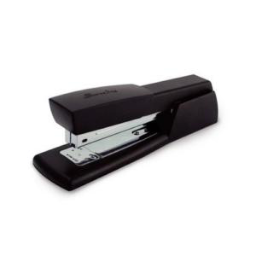}	& \includegraphics[width=0.14\textwidth]{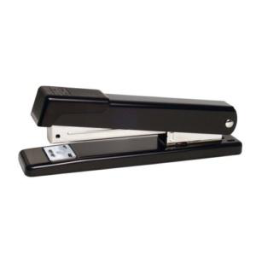} & \includegraphics[width=0.14\textwidth]{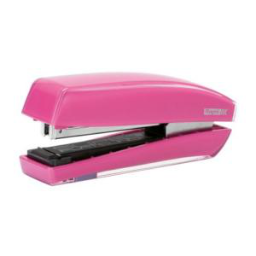} \\
		\includegraphics[width=0.14\textwidth]{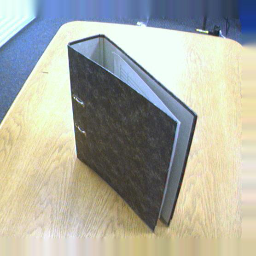} & \includegraphics[width=0.14\textwidth]{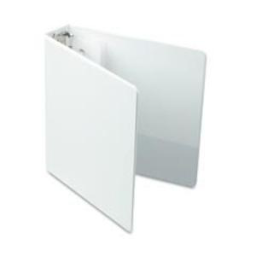} & \includegraphics[width=0.14\textwidth]{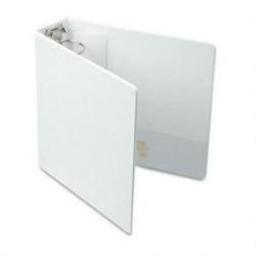} & \includegraphics[width=0.14\textwidth]{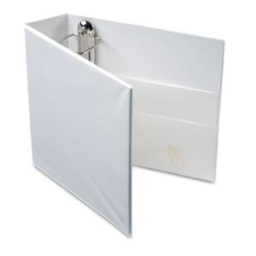}	& \includegraphics[width=0.14\textwidth]{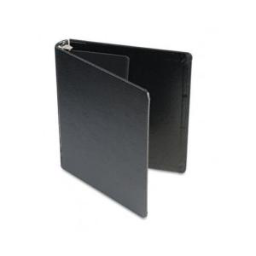} & \includegraphics[width=0.14\textwidth]{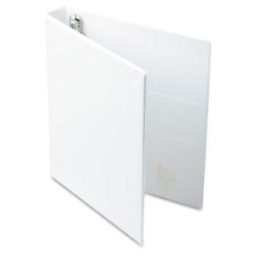} \\
		\includegraphics[width=0.14\textwidth]{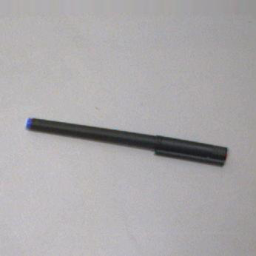} & \includegraphics[width=0.14\textwidth]{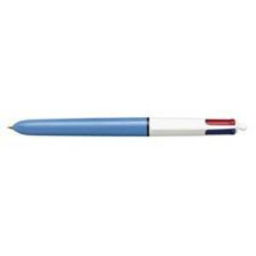} & \includegraphics[width=0.14\textwidth]{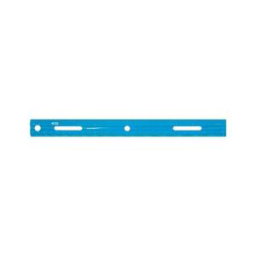} & \includegraphics[width=0.14\textwidth]{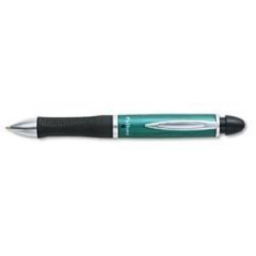}	& \includegraphics[width=0.14\textwidth]{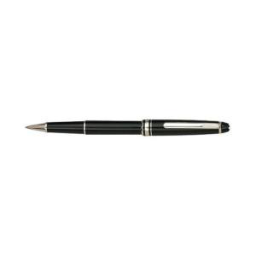} & \includegraphics[width=0.14\textwidth]{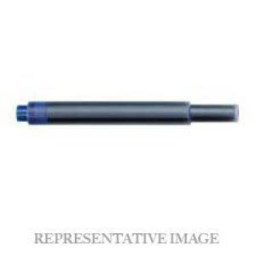} \\
	\end{tabular}
	\caption{ \label{fig:qualitativi3} Qualitative results for K-NN similarity search by \algoname{} on the \textit{Office31} dataset. Upper portion:  \textit{Amazon$\rightarrow$DSLR} scenario, lower portion:  \textit{Amazon$\rightarrow$Webcam}.}
\end{figure*}

\begin{figure*}
	\centering
	\scalebox{0.8}{
	\begin{tabular}{c|ccccc}
		\toprule
		\multicolumn{6}{c}{\textbf{Failure Cases}}\\
		\midrule
		\textbf{Query} & \textbf{1-NN} & \textbf{2-NN} & \textbf{3-NN} & \textbf{4-NN} & \textbf{5-NN} \\
		\includegraphics[width=0.14\textwidth]{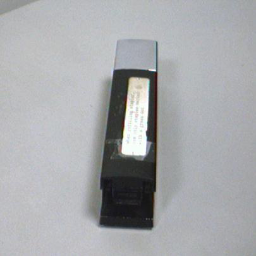} & \includegraphics[width=0.14\textwidth]{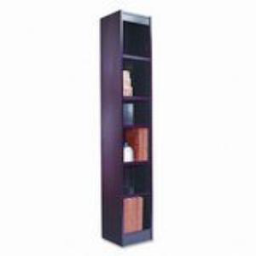} & \includegraphics[width=0.14\textwidth]{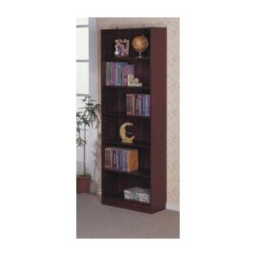} & \includegraphics[width=0.14\textwidth]{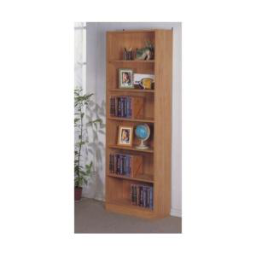}	& \includegraphics[width=0.14\textwidth]{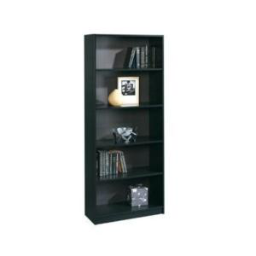} & \includegraphics[width=0.14\textwidth]{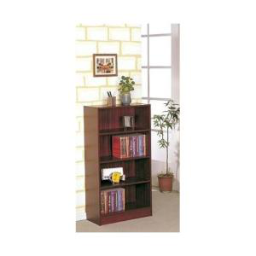} \\
		\includegraphics[width=0.14\textwidth]{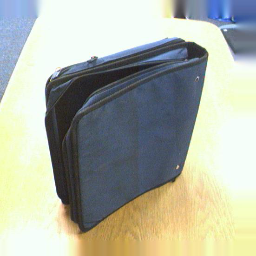} & \includegraphics[width=0.14\textwidth]{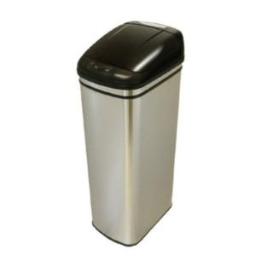} & \includegraphics[width=0.14\textwidth]{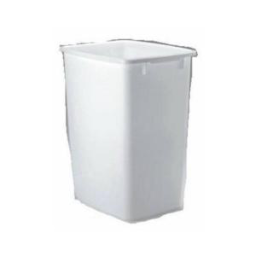} & \includegraphics[width=0.14\textwidth]{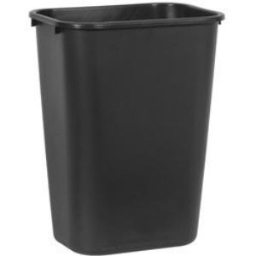}	& \includegraphics[width=0.14\textwidth]{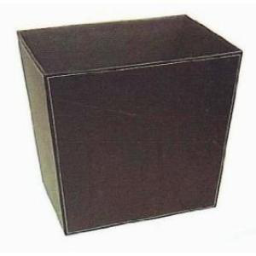} & \includegraphics[width=0.14\textwidth]{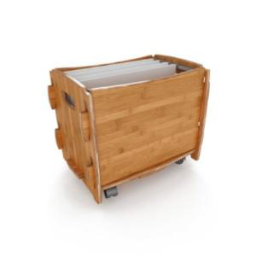} \\
		\includegraphics[width=0.14\textwidth]{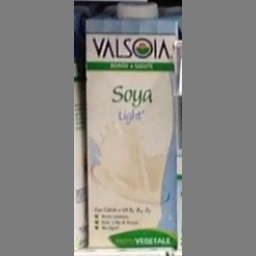} & \includegraphics[width=0.14\textwidth]{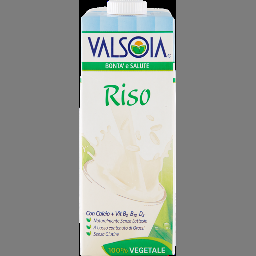} & \includegraphics[width=0.14\textwidth]{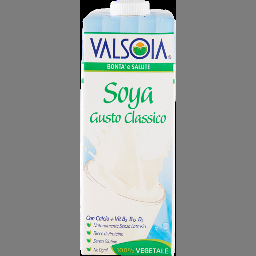} & \cellcolor{red!16!green!68!blue}\includegraphics[width=0.14\textwidth]{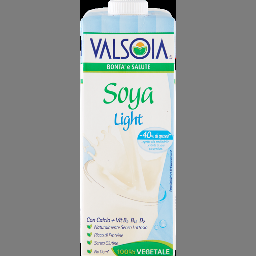}	& \includegraphics[width=0.14\textwidth]{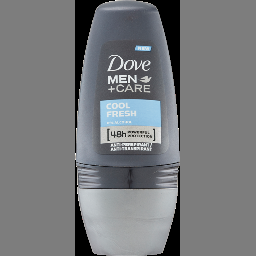} & \includegraphics[width=0.14\textwidth]{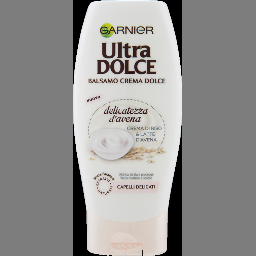} \\
		\includegraphics[width=0.14\textwidth]{qualitative_Zurigo_food_16_query_img.png} & \includegraphics[width=0.14\textwidth]{qualitative_Zurigo_food_16_k_0.png} & \includegraphics[width=0.14\textwidth]{qualitative_Zurigo_food_16_k_1.png} & \includegraphics[width=0.14\textwidth]{qualitative_Zurigo_food_16_k_2.png}	& \includegraphics[width=0.14\textwidth]{qualitative_Zurigo_food_16_k_3.png} & \includegraphics[width=0.14\textwidth]{qualitative_Zurigo_food_16_k_4.png} \\
	\end{tabular}
	}
	\caption{\label{fig:failure} Some wrong recognitions yielded by \algoname{} on the different datasets. }
\end{figure*}

\autoref{fig:qualitativi2} and \autoref{fig:qualitativi3} report some successful recognitions obtained by K-NN similarity search based on \algoname{}. The upper portion of \autoref{fig:qualitativi2}, dealing with the \textit{Grocery\_Food} scenario, shows query images for products quite hard to recognize due to both the \textit{reference} database featuring several items looking remarkably similar as well as nuisances like shadows and partial occlusions (first query) or slightly deformed products (third query). The lower portion of \autoref{fig:qualitativi2} concerns the \textit{Product8600} scenario: as clearly highlighted by the third and second query, despite differences between items belonging to the same brand and category (\eg{,} the pasta boxes in the last row) being often subtle,  \algoname{} can recognize products correctly. In \autoref{fig:qualitativi3} we report some results dealing with the \textit{Office-31} dataset which vouch how  \algoname{} can correctly retrieve images depicting objects belonging to the same category as the query. In \autoref{fig:failure}, we highlight some failure cases. In the first row \algoname{} wrongly recognizes a stapler as a bookshelf, whilst in the second a ring binder is mistaken as a trash bin. In the third row \algoname{} seems to recognize the macro class and brand of the query product though  missing the correct instance level label (\ie, the correct NN, highlighted in green, is retrieved as the 3-NN). A similar issue pertains the fourth query image: all the first 5-NNs belong to the \textit{Tea} macro class, but the correct item is not ranked among them. 

\begin{figure*}
	\centering
	\scalebox{0.8}{
	\begin{tabular}{cc|cc}
		\toprule
		\multicolumn{2}{c}{\textbf{Webcam}}&\multicolumn{2}{c}{\textbf{Grocery\_Food}}\\
		\midrule
		\textbf{Original}&\textbf{Generated}&\textbf{Original}&\textbf{Generated}\\
		\includegraphics[width=0.20\textwidth]{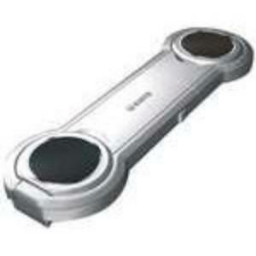} & \includegraphics[width=0.20\textwidth]{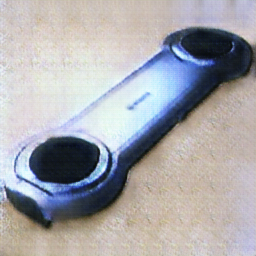} & \includegraphics[width=0.20\textwidth]{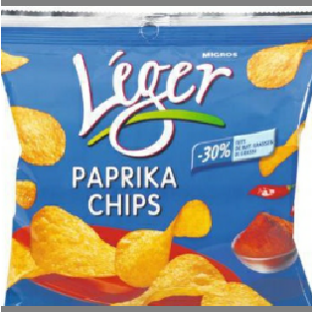} & \includegraphics[width=0.20\textwidth]{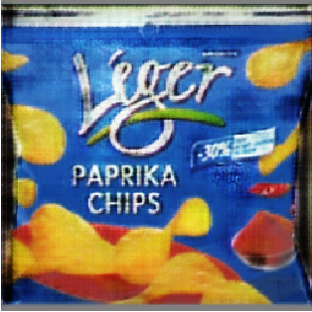} \\
		\includegraphics[width=0.20\textwidth]{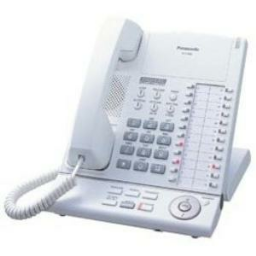} & \includegraphics[width=0.20\textwidth]{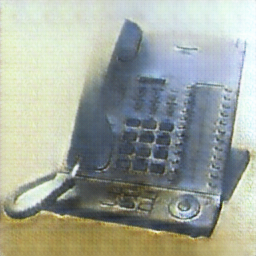} & \includegraphics[width=0.20\textwidth]{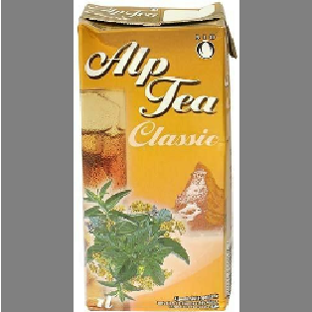} & \includegraphics[width=0.20\textwidth]{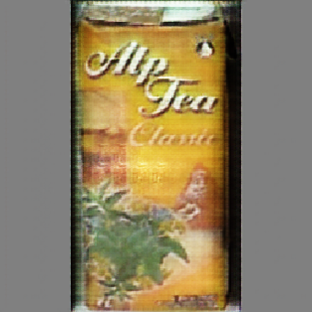} \\
		\includegraphics[width=0.20\textwidth]{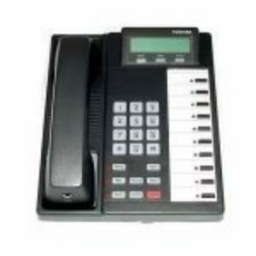} & \includegraphics[width=0.20\textwidth]{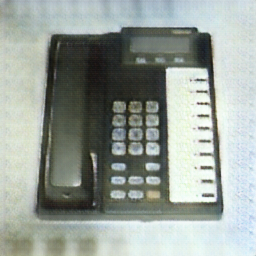} & \includegraphics[width=0.20\textwidth]{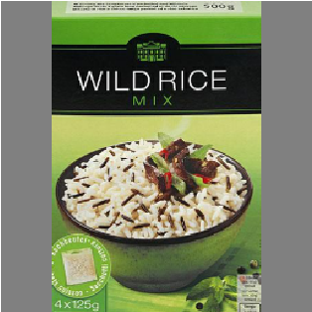} & \includegraphics[width=0.20\textwidth]{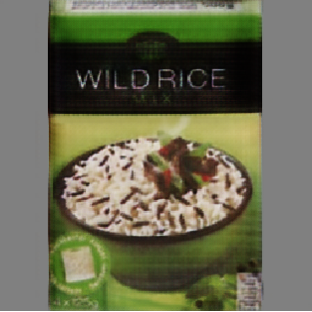} \\
	\end{tabular}
	}
	\caption{\label{fig:gan} Images generated by the GAN trained jointly with the embedding network in \algoname{} for the \textit{Office31-Amazon$\rightarrow$Webcam} (left) and \textit{Grocery\_Food} (right) scenarios. Columns labeled as \textit{Original} depict images provided as input to the \textit{Generator} while those labeled as \textit{Generated} show the corresponding outputs. }
\end{figure*}

Finally, in \autoref{fig:gan} we show some training images generated by our GAN framework to pursue domain invariance (\autoref{ss:dom_inv}).  It is worth observing how the training samples created by our GAN seem to preserve both the overall structure and details of the input images, which is very important when addressing instance level recognition between  many similarly looking items,  while modifying significantly the brightness and colors and injecting some blur, thereby realizing a domain translation between the input and output images.

\section{Conclusion}
The experimental results demonstrate how our proposed hierarchical modification to the triplet ranking loss is effective in learning  embedding functions that generalize better to unseen data. Moreover, the integration of a GAN at training time, so as to obtain the whole \algoname{} architecture, turns out a very effective approach in scenarios where only few/one samples per class are available at training time. Indeed, \algoname{} allows for learning a representation not only robust to domain shift but also  better all around due to the \textit{Generator} network effectively producing potentially infinite hard training samples. \algoname{} was designed to solve the task of Grocery Product recognition, in which it can deliver  remarkably good performance. Nonetheless, experiments in different scenarios suggest that our architecture may be effectively deployed in settings that features challenges such as few training samples per class, different domains between train and test data and a taxonomy among classes to be recognized. The results of the ablation study of \autoref{ss:ablation} show clearly  how the main improvement in performance is provided by the original introduction of a GAN network trained end-to-end together with the embedding network in order to augment the training set. In the future we would like to further investigate on this novel concept through different combination of embedding losses and GAN architectures. For example, a possible variant of \algoname{} may concern a GAN that would operate at the feature embedding level rather than at image level, thereby hallucinating embedding vectors from the domain $\mathcal{B}$ given those from domain $\mathcal{A}$, with these vectors amenable to compute any kind of embedding loss.    

Finally, in this paper we have proposed an architecture to recognize a product item extracted from a shelf image, though we did not address how to actually detect the individual product items within such an image. Recent works like \cite{Qiao_2017_ICCV} have shown how a region proposal CNN can be successfully trained to extract bounding boxes surrounding grocery products from an image featuring the whole shelf. Thus, \cite{Qiao_2017_ICCV} and \algoname{} may be combined  effortlessly in a complete pipeline that given a single shelf image and one reference image for each product can detect and recognize the displayed items.

\section*{Acknowledgments}
We would like to thank \textit{Centro Studi s.r.l.} for funding this research project.

\bibliographystyle{ieee}
\bibliography{embedding.bib}

\end{document}